\documentclass[conference]{IEEEtran}
\IEEEoverridecommandlockouts
\usepackage{cite}
\usepackage{amsmath,amssymb,amsfonts}
\usepackage{algorithmic}
\usepackage{graphicx}
\usepackage{textcomp}
\usepackage{xcolor}

\usepackage{enumitem}
\usepackage[switch]{lineno}
\usepackage{bbding} 
\usepackage{amsmath}
\usepackage{amssymb}
\usepackage{balance} 
\usepackage{url}
\usepackage{algorithm}
\usepackage{algorithmic}
\usepackage{graphicx}
\usepackage{multirow} 
\usepackage{booktabs}
\usepackage{makecell}
\usepackage{colortbl}
\definecolor{gray}{RGB}{222,222,222}
\usepackage{subfigure}
\def\BibTeX{{\rm B\kern-.05em{\sc i\kern-.025em b}\kern-.08em
    T\kern-.1667em\lower.7ex\hbox{E}\kern-.125emX}}
    
\makeatletter
\newcommand{\linebreakand}{%
  \end{@IEEEauthorhalign}
  \hfill\mbox{}\par
  \mbox{}\hfill\begin{@IEEEauthorhalign}
}
\makeatother

\begin{document}

\title{Cross-domain Federated Object Detection}

\author{
\IEEEauthorblockN{Shangchao Su, Bin Li$^{*}$\thanks{$^{*}$Corresponding author.}, Chengzhi Zhang, Mingzhao Yang, Xiangyang Xue$^{*}$}
\IEEEauthorblockA{
\textit{School of Computer Science, Fudan University}\\
\{scsu20, libin, zhangcz20, mzyang20, xyxue\}@fudan.edu.cn}
}

 
\maketitle

\begin{abstract}
Detection models trained by one party (including server) may face severe performance degradation when distributed to other users (clients). Federated learning can enable multi-party collaborative learning without leaking client data. In this paper, we focus on a special cross-domain scenario in which the server has large-scale labeled data and multiple clients only have a small amount of labeled data; meanwhile, there exist differences in data distributions among the clients. In this case, traditional federated learning methods can't help a client learn both the global knowledge of all participants and its own unique knowledge. To make up for this limitation, we propose a cross-domain federated object detection framework, named FedOD. The proposed framework first performs the federated training to obtain a public global aggregated model through multi-teacher distillation, and sends the aggregated model back to each client for fine-tuning its personalized local model. After a few rounds of communication, on each client we can perform weighted ensemble inference on the public global model and the personalized local model. We establish a federated object detection dataset which has significant background differences and instance differences based on multiple public autonomous driving datasets, and then conduct extensive experiments on the dataset. The experimental results validate the effectiveness of the proposed method.
\end{abstract}


	\section{Introduction}

In some real-world scenarios, a centrally trained detection model (usually trained on a server) will be sent to multiple users (clients). However, the clients are often in different domains. For example, in autonomous driving scenarios, distribution discrepancy in weather, lighting, and vehicle type among different clients may all affect the prediction performance of the model. Therefore, it is necessary to perform collaborative learning on the datasets of the server and multiple clients. The trained model needs to learn new knowledge in the client domain while still preserving server's knowledge. Moreover, for the purpose of privacy protection, during the collaborative learning, the server cannot collect the raw datasets from the clients, which brings a great challenge.

Federated learning (FL) is a technology that has the potential to address this challenge. FL aims to allow each participant to perform joint machine learning without sharing its local data. The original FL~\cite{mcmahan2017communication} needs to learn a single global model to handle the requirements of multiple clients, i.e., one model fits all. In each round of communication, the server sends a global model to each client. The client uses its local data to update the model and sends the trained model (local model) back to the server to perform parameter averaging, and the server uses the averaged model as a new aggregated model. After multiple rounds of communication, the converged aggregated model is able to learn the knowledge from multiple clients. Some studies~\cite{DBLP:conf/iclr/LiHYWZ20,DBLP:journals/corr/abs-1910-14425,SU2023} have shown that FedAvg~\cite{mcmahan2017communication} can converge to good performance when the datasets of multiple clients are independently identically distributed (iid); but the performance is likely to drop significantly when the data is non-iid, i.e., the data for each client comes from different distributions. Unlike the original FL which aims to improve the performance of the global model, personalized federated learning (pFL) learns a personalized model for each client, thereby improving the performance of all client models and better handling non-iid challenges. For example, FedRep~\cite{DBLP:conf/icml/CollinsHMS21} enables each client to compute a personalized classifier and a globally shared feature extractor. FedBN~\cite{li2021fedbn} provides personalized batch normalization (BN) layers for different clients. These personalized methods can improve the traditional FL in non-iid scenarios to a certain extent.

		\begin{figure}[t]
		\centering
		{\includegraphics[width=0.88\linewidth]{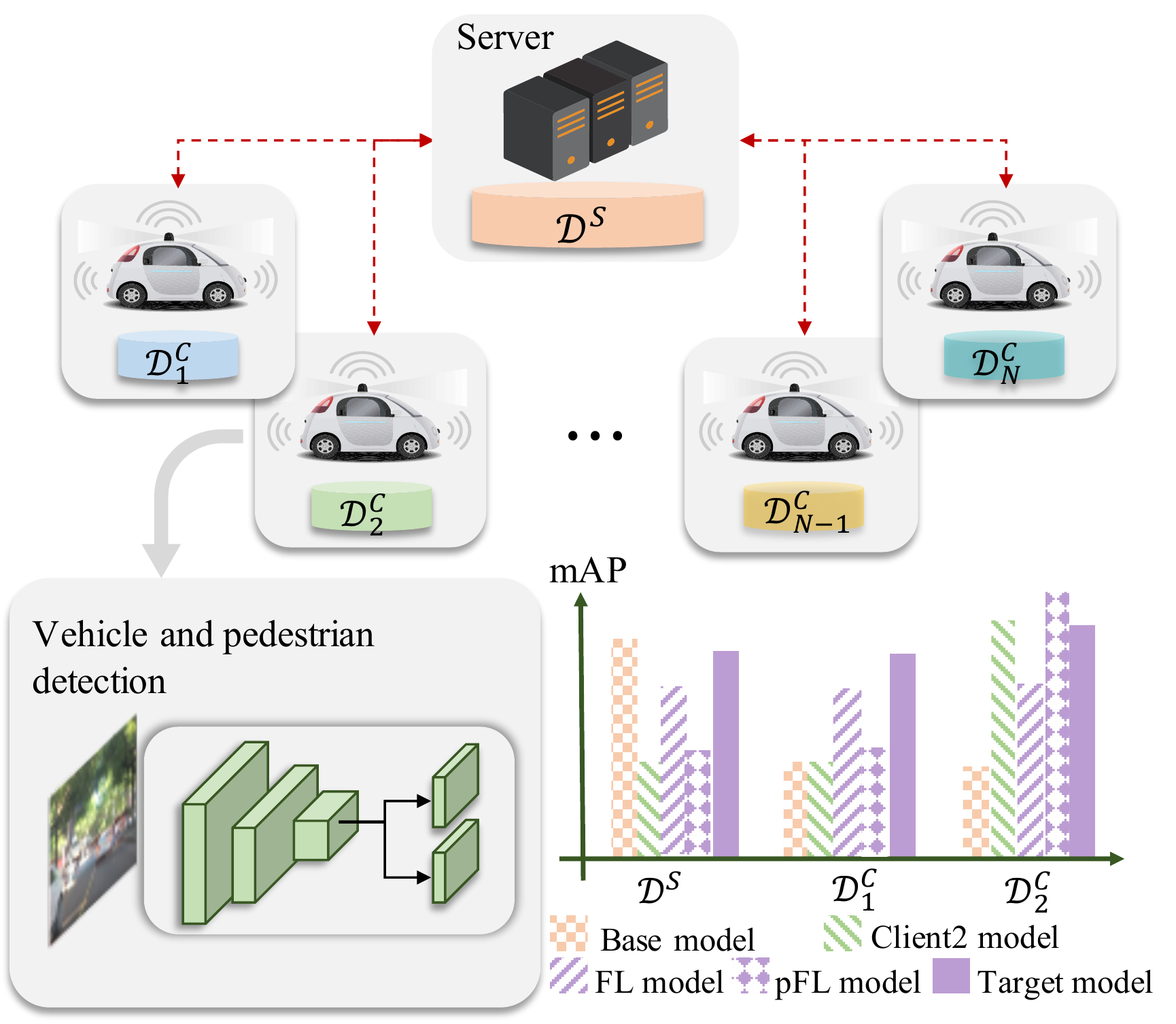}}
		\caption{The problem scenario. The server has a large amount of labeled data, each client has only a small amount of labeled data, and the data distribution of each client is different from one another. The base model trained on $\mathcal{D}^S$ suffers severe performance degradation on client-side datasets. 
		}
		\label{fig:task} 
	\end{figure}
	

However, in some practical scenarios, such as the server-client collaborative autonomous driving scenario (Figure~\ref{fig:task}), there are usually a server with a large-scale labeled dataset (more than ten times the data of the client) and multiple clients, each of which only has a small amount of labeled data. There is a distinct domain shift between each client and the server. In such a scenario, on one hand, the original FL may be affected by data imbalance and the non-iid issue, making personalized knowledge difficult to learn. On the other hand, while pFL is proposed to improve the client performance, it results in insufficient learning of the global domain (the union of the server and all clients). See Figure~\ref{fig:task}, the original FL (`FL model') focuses on the performance of the global domain, which can hardly obtain high performance on Client 2. pFL (`pFL model') focuses on personalized knowledge, but has lower performance on the server and Client 1. In the cross-domain federated object detection scenario considered in this paper, we aim to achieve this target, that for each client it is able to learn both its own personalized knowledge and the knowledge of most other domains.
	


To achieve this goal, in this paper, we propose a distillation-based federated learning framework for cross-domain Federated Object Detection (FedOD). Each client maintains two models, a public global model and a personalized local model. The public model is the aggregated model received from the server, the personalized model is a model trained based on the aggregated model and the local dataset. In each round of communication, each client's personalized model is sent to the server, and then the server data is used to perform multi-teacher distillation on these personalized models to obtain the aggregated model. By mimicking the intermediate representations of the personalized models, the aggregated model can distill more global knowledge through using the server data. After a few rounds of communication, the public global model and the personalized local model are used to perform ensemble inference. For moderating the importance of the public global model and the personalized local model, we adopt the weighted bounding boxes (for simplicity, we use `bboxes' in the following) fusion strategy. In this way, each client can maintain both global knowledge and local personalized knowledge. Note that the performance of the ensemble model in the global domain can outperform a single model with the same parameter scale.

In order to validate the proposed framework, we construct a collaborative learning dataset for vehicle pedestrian detection based on three public datasets. We extract annotated images from BDD100K~\cite{yu2020bdd100k} to build the dataset for server, use the annotated images from SODA10M~\cite{han2021soda10m} to construct Client 1 and Client 2, and the NuScenes~\cite{caesar2020nuscenes} dataset for Client 3 and Client 4. In addition to the distribution differences that these datasets themselves have (regions, lighting, street types, shooting equipment, etc.), in order to introduce differences into instance class distributions, we choose some specific types of vehicles (e.g., motorcycles and trucks) to keep them only in the two NuScenes clients and remove them from the server and the other clients.

Our contributions are summarized as follows: 1) We propose a new realistic federated learning scenario, which is the first attempt to consider server-client collaborative object detection, in which, neither the original FL nor the pFL methods can fully meet the requirements. 2) We propose a federated object detection (FedOD) framework that employs multi-teacher distillation and weighted bboxes fusion to enable each client to simultaneously obtain global knowledge and personalized knowledge. 3) We establish a federated learning dataset based on public datasets and conduct extensive experiments to validate the effectiveness of our proposed method.


	\section{Method}\label{sec:method}
	
	\subsection{Problem Setting}
Suppose we have a server and $N$ clients. The $i$-th client has its own data distribution $\mathcal{P}_i$, and its dataset $\mathcal{D}^C_i$ is collected from $\mathcal{P}_i$; while the server has dataset $\mathcal{D}^S$ collected from distribution $\mathcal{P}^S$, where $|\mathcal{D}^S| \gg |\mathcal{D}_i^C|$ for $ i = 1,\cdots, N$. Note that there exist obvious differences among different data distributions. 

The original federated learning is to learn a global model that works well for all clients, a classic description is to minimize the objective $\min _{\mathbf{w} \in \mathbb{R}^{d}} \frac{1}{N} \sum_{i=1}^{N} L_{i}(\mathbf{w})$,
where $L_{i}(\mathbf{w})\triangleq \mathbb{E}_{\mathbf{x} \sim \mathcal{P}_{i}}\left[l(\mathbf{w} ; \mathbf{x})\right]$, $l$ is the loss function. Based on this objective, in each round, FedAvg~\cite{mcmahan2017communication} samples $m$ clients and collects the trained models from these clients. Then the aggregated model is  $\sum_{i=1}^m \lambda_i \mathbf{w}_i$, where $\lambda_i$ is the importance of the $i$-th model, usually determined by the amount of data on the client. In addition, $\sum_i \lambda_i=1$.

Personalized federated learning is to minimize the following objective function to assign parameter $\mathbf{w}_i$ to the $i$-th client: $ \min _{\{\mathbf{w}_1,\cdots,\mathbf{w}_N\}} \frac{1}{N} \sum_{i=1}^{N} L_{i}(\mathbf{w}_i)$.

In this paper, we consider a new yet realistic federated learning setting: A large amount of labeled data exists on the server, so it is crucial to make good use of the knowledge on the server side. Our goal is that the client can learn new personalized knowledge from each client while still preserving the knowledge of the global domain. To this end, the following objective is adopted: 
\begin{linenomath}
\begin{align}
    \min _{\{\mathbf{w}_1,\cdots,\mathbf{w}_N\}} \frac{1}{N} \sum_{i=1}^{N} L_U(\mathbf{w}_i)
    \label{purpose}
\end{align}
\end{linenomath}
where $L_U\triangleq \mathbb{E}_{\mathbf{x} \sim \mathcal{P}^S\cup\mathcal{P}_1\cup\cdots\cup\mathcal{P}_N}\left[l(\mathbf{w} ; \mathbf{x})\right]$. $L_U$ evaluates the performance of the local model in both its own domain and in other domains. Note that when testing for a certain client, due to the large scale of the server data, $\mathbb{E}_{\mathbf{x} \sim \mathcal{D}^S\cup\mathcal{D}^C_1\cup\cdots\mathcal{D}^C_N}\left[l(\mathbf{w} ; \mathbf{x})\right]$ will be dominated by the server data. Even if the performance on the client testset is greatly improved, a small decline on the server dataset may lead to an observable decline in the performance of the overall union set. Thus the learning effect of personalized knowledge will be covered up. Therefore, in the experimental part, we will show the performance of the model in both the global domain and the local domain.

		\begin{figure}[t]
		\centering
		{\includegraphics[width=0.88\linewidth]{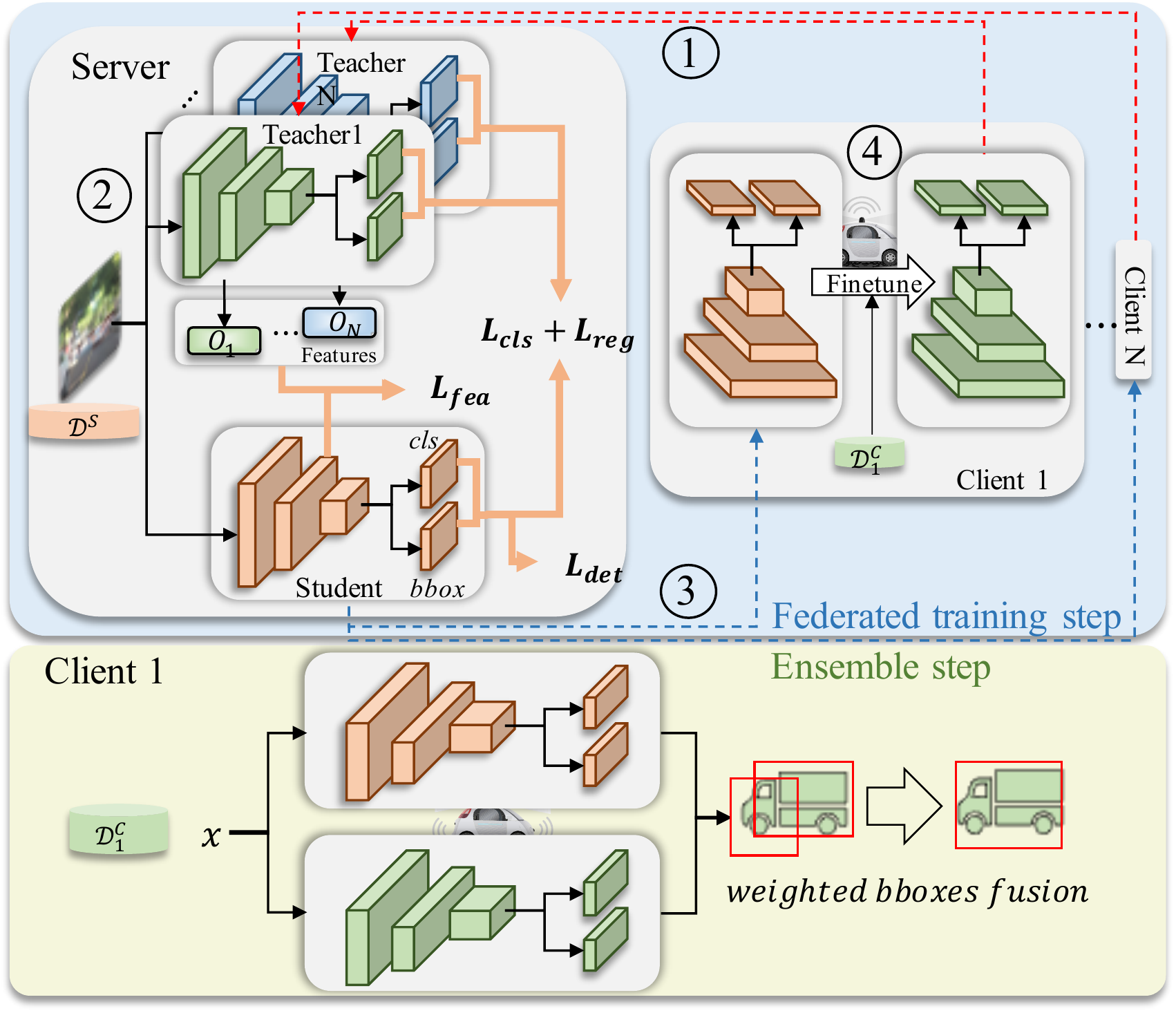}}
		\caption{Overview of the proposed framework for cross-domain federated object detection. We first conduct a few rounds of communication. Second, we apply weighted bboxes fusion to ensemble the outputs of these two models.
		}
		\label{fig:framework} 
	\end{figure}
	
\subsection{FedOD}
\textbf{Overview.} The overall illustration is shown in Figure~\ref{fig:framework}. Suppose we have a base model $\mathbf{w}_b$ trained in the server domain. If $\mathbf{w}_b$ is sent directly to the clients, the performance of $\mathbf{w}_b$ on the client-side will be greatly reduced due to the difference among data distributions of the clients and the server. Our goal is to develop a better model from $\mathbf{w}_b$ within few communication rounds to achieve Eq.\ref{purpose}. The proposed framework is illustrated in Figure~\ref{fig:framework}: The first step is federated training. In this step, each client always maintains two models, a global public model and a local personalized model. In each communication round, each client sends its personalized model to the server (\textcircled{1}), and the server employs multi-teacher distillation to aggregate these personalized models (\textcircled{2}). The aggregated model is then sent back to the client as a global public model (\textcircled{3}), and the client uses its local data to update the public model (in this paper, we fine-tune the public model) to obtain a personalized model (\textcircled{4}). The second step is ensemble. In order to solve the dilemma that the client model can not take into account the global and local knowledge in the traditional FL setting, 
we adopt a new idea to improve the client's ability of learning global and local knowledge through ensembles. We make use of the two detection models maintained by each client to ensemble, i.e., perform weighted bounding boxes (bboxes) fusion on the outputs of the global public model and the local personalized model.

\begin{algorithm}[t]
\caption{FedOD}
\begin{algorithmic}
\STATE {\textbf{The Federated Training Step:}}
  \STATE Initialize global model $\mathbf{w}_g \leftarrow \mathbf{w}_b$
  \FOR{each round $t = 1, 2, \dots$}
     \STATE $S_t \leftarrow$ Randomly select $n$ clients 
     \FOR{each client $i \in S_t$ \textbf{in parallel}}
      \STATE $\mathbf{w}_i \leftarrow \text{ClientTrain}(\mathcal{D}^C_i, \mathbf{w}_g)$ 
     \ENDFOR
     \STATE Initialize the student model $\mathbf{w}_g$ by Eq.2.
     \STATE Update $\mathbf{w}_g$ by Eq.3 using $\{\mathbf{w}_i\}_{i=1}^n$ as teachers.
  \ENDFOR
\STATE {\textbf{The Ensemble Step:}}
\FOR{each client $i = 1, 2, \dots$}
 \STATE Get the latest $\mathbf{w}_g$ from the server
 \STATE Finetune $\mathbf{w}_g$ with local dataset $\mathcal{D}^C_i$
 \STATE Ensemble $\mathbf{w}_g$ and $\mathbf{w}_i$ as the final model of the $i$-th client.
\ENDFOR
\end{algorithmic}
\label{alg}
\end{algorithm}

\textbf{The Federated Training Step.}
We observe that when there are obvious differences in the data distributions of clients, the public global aggregated model itself $\mathbf{w}_g$ seems unable to simultaneously extract the knowledge from the server and multiple clients within a few number of communication rounds, so we resort to introducing an additional personalized model $\mathbf{w}_i$ on the $i$-th client at the same time. In the federated training step, the global model and the personalized models of all the clients are updated every round. The global model is aimed to learn as much knowledge of the global domain as possible. The personalized model is expected to learn more personalized knowledge based on the continuously improved global model, because under the assumption of pFL, global knowledge is conducive to the learning of personalized knowledge. By alternately updating these two models, the proposed framework can acquire more comprehensive global and personalized knowledge for the subsequent ensemble step. Note that $\mathbf{w}_g$ is not necessary to perform the best in all client domains, but needs to be able to capture the knowledge of most parties involved, that is, $\mathbf{w}_g$ needs to outperform the base model $\mathbf{w}_b$ in the global domain $\mathcal{P}^S\cup\mathcal{P}_1\cup\cdots\cup\mathcal{P}_N$.

During federated training, each client first sends its personalized model $\mathbf{w}_i$ to the server. Next, the server is responsible for aggregating these local models. The traditional FL methods only conduct simple parameter averaging in the server, and do not fully utilize the server data. Therefore, to significantly improve the model aggregation efficiency, we employ the multi-teacher distillation algorithm for model aggregation on the server. Before distillation, the student model is initialized by averaging the aggregated model of the previous round $\mathbf{w}_g$ and the personalized models $\mathbf{w}_i$:
\begin{align}
\label{kd}
    \mathbf{w}_g\leftarrow \frac{1}{N+1}\left[\mathbf{w}_g + \sum_{i=1}^N \mathbf{w}_i\right]
\end{align}
The feature extraction part of the detection model includes feature maps of multiple scales, we add the channel-wise distillation loss~\cite{shu2021channel} to the output feature maps of each scale respectively. Suppose that the feature map of the $l$-th scale of the $i$-th client (teacher) model is $O^{C,l}_i$ and the feature map of the student model is $O^{S,l}$, we first obtain an attention map by applying softmax to these features: $ \mathcal{A}(O_{k,j})=\frac{exp(\frac{O_{k,j}}{\mathcal{T}})}{\sum_{j=1}^{W*H} exp(\frac{O_{k,j}}{\mathcal{T}})}$
where $k$ indexes the channel, $W$ and $H$ are the width and height of the feature map, respectively; $\mathcal{T}$ is the temperature parameter. Then the KL divergence is used to measure the distance between the feature maps of the student and each teacher:
\begin{linenomath}
\begin{align}
\nonumber
    KL\left(O^{S,l}, O^{C,l}_i\right)=\frac{\mathcal{T}^{2}}{K} \sum_{k=1}^{K} \sum_{j=1}^{W_l \cdot H_l} \mathcal{A}\left(O_{k, j}^{S,l}\right) \cdot \log \frac{\mathcal{A}\left(O_{k, j}^{S,l}\right)}{\mathcal{A}\left(O_{i,k,j}^{C,l}\right)}
\end{align}
\end{linenomath}
The overall feature loss is $\mathcal{L}_{fea} =\frac{1}{N} \sum_l \sum_i KL(O^{S,l}, O^{C,l}_i)$.
Note that a more common multi-teacher distillation method is to calculate the KL divergence after averaging the features, that is, use $KL\left(O^S,\frac{1}{N}\sum_i^N O^C_i\right)$. However, under the setting of this paper, the training data of each teacher model is quite different. If the features are directly averaged, their personalized information will be lost.

For detection models, such as RetinaNet~\cite{lin2017focal}, the output of the classification head can also be regarded as a feature map, so $\mathcal{L}_{cls}$ has the same formula as $\mathcal{L}_{fea}$. For the regression head, we use the L2 distance as $\mathcal{L}_{reg}$. The overall loss for the distillation is:
\begin{linenomath}
\begin{align}
\label{distill}
    \mathcal{L}_{distillation} = \lambda_1*\mathcal{L}_{fea}+\lambda_2*\mathcal{L}_{cls}+\lambda_3*\mathcal{L}_{reg}
\end{align}
\end{linenomath}
Furthermore, the original detection loss $\mathcal{L}_{det}$ is still applied to the student model. After distillation, the student model, as a new aggregated model $\mathbf{w}_g$, is sent to the clients. The details of the proposed FedOD is shown in Algorithm~\ref{alg}.

\textbf{The Ensemble Step.}
After the previous step, for each client, we obtain an aggregated model $\mathbf{w}_g$, which has better generalization performance compared with the base model $\mathbf{w}_b$. Also we have a personalized model which fully excavates the personalized knowledge of the client. However, the unified aggregated model itself cannot adapt to the personalized knowledge of all clients, and the personalized model itself cannot well handle the samples from other domains. Therefore, we consider keeping these two models at the same time and ensembling their outputs. Although we introduce an additional model compared to the pFL setting, it will have better performance under the same number of model parameters. We will show this in the experiment section.
	
The key to ensemble the detection models is how to filter the predicted bboxes. In order to ensemble the models trained from the same domain, a simple method is to use traditional non-maximum suppression (NMS) algorithm. However, under the setting of this paper, the public global model and the personalized local model are not trained from the same domain. Due to domain shift, these two models are likely to have different understandings of the same object, we cannot simply decide which box to discard and which to keep. In this case, standard NMS methods have the possibility to ignore valid predictions, we can't simply drop one and keep the other, but should fuse them. See supplementary materials for more analysis. To this end, we adopt weighted boxes fusion (WBF)~\cite{solovyev2021weighted} algorithm, which uses all the bboxes. 
The idea of WBF is simple. First, put the output bboxes of the two models into set $\mathcal{B}$, then cluster the predictions in $\mathcal{B}$ by the preset IoU threshold and fuse the bboxes for each cluster. Suppose a certain cluster has $T$ bboxes, and the confidence of the $i$-th bbox is $C_i$, the coordinate vector is $\mathbf{b}_i$, then the fusion rule is: $C\leftarrow \frac{\min{(T,M)}}{M*T} \sum_{i=1}^T C_i,\,\,\, \mathbf{b} \leftarrow \frac{\sum_{i=1}^T C_i*\mathbf{b}_i}{\sum_{i=1}^T C_i}$, 
where $M$ is the number of the models to be ensembled, in this paper, $M=2$. $C$ and $\mathbf{b}$ will be the new confidence and coordinate of the fused bbox. Weighted bboxes fusion makes use of all bboxes at the same time, so it is especially suitable for the ensemble of cross-domain models in this paper. In subsequent experiments, we will compare different ensemble methods.

\begin{table*}[tb]
\centering
\caption{Performance comparison. $\mathbf{w}_b$ is the base model trained only with the server data; $\mathbf{w}^1_g$ is a model aggregated by $\{\mathbf{w}^1_i\}_{i=1}^N$; $\mathbf{w}^2_i$ is the model fine-tuned from $\mathbf{w}^1_g$ using the client datasets; $E(\cdot,\cdot)$ denotes the ensemble model. }
\resizebox{0.98\linewidth}{!}{%
\begin{tabular}{cc|c|ccc|ccc|ccc|ccc}
\Xhline{1.2pt}
\multicolumn{2}{c|}{\multirow{4}{*}{\textbf{}}} & \multirow{4}{*}{\textbf{\begin{tabular}[c]{@{}c@{}}Ideal\\  model\end{tabular}}} & \multicolumn{3}{c|}{\multirow{2}{*}{\textbf{Without Joint Learning}}} & \multicolumn{3}{c|}{\multirow{2}{*}{\textbf{FedOD 1-th round}}} & \multicolumn{3}{c|}{\multirow{2}{*}{\textbf{FedOD 2-th round}}} & \multicolumn{3}{c}{\multirow{2}{*}{\textbf{FedOD 3-th round}}} \\
\multicolumn{2}{c|}{} &  & \multicolumn{3}{c|}{} & \multicolumn{3}{c|}{} & \multicolumn{3}{c|}{} & \multicolumn{3}{c}{} \\ \cline{4-15} 
\multicolumn{2}{c|}{} &  & \multirow{2}{*}{\textbf{\begin{tabular}[c]{@{}c@{}}Baseline1 \\ $\mathbf{w}_b$\end{tabular}}} & \multirow{2}{*}{\textbf{$\mathbf{w}^1_i$}} & \multirow{2}{*}{\textbf{\begin{tabular}[c]{@{}c@{}}Baseline2\\  $E(\mathbf{w}^1_i,\mathbf{w}_b)$\end{tabular}}} & \multirow{2}{*}{\textbf{$\mathbf{w}^1_g$}} & \multirow{2}{*}{\textbf{$\mathbf{w}^2_i$}} & \multirow{2}{*}{\textbf{$E(\mathbf{w}^2_i,\mathbf{w}^1_g)$}} & \multirow{2}{*}{\textbf{$\mathbf{w}^2_g$}} & \multirow{2}{*}{\textbf{$\mathbf{w}^3_i$}} & \multirow{2}{*}{\textbf{$E(\mathbf{w}^3_i,\mathbf{w}^2_g)$}} & \multirow{2}{*}{\textbf{$\mathbf{w}^3_g$}} & \multirow{2}{*}{\textbf{$\mathbf{w}^4_i$}} & \multirow{2}{*}{\textbf{$E(\mathbf{w}^4_i,\mathbf{w}^3_g)$}} \\
\multicolumn{2}{c|}{} &  &  &  &  &  &  &  &  &  &  &  &  &  \\ \Xhline{1pt}
\multicolumn{2}{c|}{$A_s$} & \textit{26.90} & \textit{\textbf{25.50}} & \textit{18.95} & \textit{24.23} & \textit{24.50} & \textit{18.73} & \textit{23.13} & \textit{24.20} & \textit{18.43} & \textit{22.90} & \textit{24.00} & \textit{18.18} & \textit{22.53} \\  \hline
 \multicolumn{2}{c|}{$A_p$} & \textit{42.83} & \textit{21.25} & \textit{39.83} & \textit{38.65} & \textit{32.95} & \textit{42.00} & \textit{41.18} & \textit{35.20} & \textit{42.38} & \textit{41.93} & \textit{35.78} & \textit{\textbf{42.90}} & \textit{42.45} \\
 \hline
\multicolumn{2}{c|}{$A_u$} & \textit{41.30} & \textit{22.60} & \textit{29.33} & \textit{31.80} & \textit{32.50} & \textit{30.13} & \textit{33.93} & \textit{33.40} & \textit{30.25} & \textit{34.53} & \textit{33.90} & \textit{30.35} & \textit{\textbf{34.88}} \\ \hline
\multicolumn{1}{c|}{\multirow{3}{*}{$A_{com}$}} & $\alpha=0.1$ & \textit{41.45} & \textit{22.47} & \textit{30.38} & \textit{32.49} & \textit{32.55} & \textit{31.31} & \textit{34.65} & \textit{33.58} & \textit{31.46} & \textit{35.27} & \textit{34.09} & \textit{31.61} & \textit{\textbf{35.63}} \\ \cline{2-15} 
\multicolumn{1}{c|}{} & $\alpha=0.3$ & \textit{41.76} & \textit{22.20} & \textit{32.48} & \textit{33.86} & \textit{32.64} & \textit{33.69} & \textit{36.10} & \textit{33.94} & \textit{33.89} & \textit{36.75} & \textit{34.46} & \textit{34.12} & \textit{\textbf{37.15}} \\ \cline{2-15} 
\multicolumn{1}{c|}{} & $\alpha=0.5$ & \textit{42.06} & \textit{21.93} & \textit{34.58} & \textit{35.23} & \textit{32.73} & \textit{36.06} & \textit{37.55} & \textit{34.30} & \textit{36.31} & \textit{38.23} & \textit{34.84} & \textit{36.63} & \textit{\textbf{38.66}} \\ \Xhline{1.2pt}
\end{tabular}%
}
\label{tab:baseline}
\end{table*}

	\section{Experiments}
	In this section, we start by setting up our experiments, and then introduce the construction of the dataset, report and analyze the experimental results.

	\textbf{Implementation Details.} 
	We implement our framework using MMDetection~\cite{DBLP:journals/corr/abs-1906-07155}. The detection model is RetinaNet~\cite{lin2017focal}.
	The base model $\mathbf{w}_b$ on the server is initialized by COCO2014~\cite{lin2014microsoft} pretrained parameters. We use four NVIDIA GeForce RTX 3090 GPUs to do federated training. By default, we set the batch size $B=16$, learning rate $lr=0.01$, the local training epoch to be 12, and the distillation training epoch to be 5.
	
	\textbf{Evaluation Metrics.}
	We use mean Average Precision (mAP) to evaluate the detection performance. Assume that after the federated learning step, the performance of each client on its own testset is $r^p_i$, on the server testset is $r^s_i$, and the performance on the global testset (the union of the server testset and all client testsets) is $r^u_i$. Specifically, $r^u_i$ measures the global generalization ability; $r^p_i$ measures the ability to learn the local personalized knowledge; $r^s_i$ measures the degree of forgetting of the initial knowledge on the server. In short, we finally report the following four indicators: $A_s = \frac{1}{N}\sum^N_{i=1}r^s_i$, $A_p = \frac{1}{N}\sum^N_{i=1}r^p_i$, $A_u = \frac{1}{N}\sum^N_{i=1}r^u_i$ and $A_{com} = \frac{1}{N}\sum^N_{i=1} \alpha*r^p_i+(1-\alpha)*r^u_i$,  where $\alpha$ is a trade-off hyperparameter. Note that for the real dataset, the scale of server data is large, which makes $A_u$ unable to reflect the learning ability of personalized knowledge. Therefore, we should use the weighted sum of $A_u$ and $A_p$. We report $A_{com}$ with different $\alpha$ in Table~\ref{tab:baseline} to observe the effect of $\alpha$.

	\textbf{Compared Methods.}
	Two baselines and three competitors are compared as follows. 1) \textbf{Baseline1}: the base model $\mathbf{w}_b$ of the server. This represents the performance of doing nothing but using the base model as the local model for each client.
	2) \textbf{Baseline2}: to show the effectiveness of federated learning, we directly ensemble the base model and the first round finetuned model on the client side, as baseline2 ($E(\mathbf{w}^1_i,\mathbf{w}_b)$), the final ensemble results of FedOD should be better than baseline2. 3)  \textbf{FedAvg}~\cite{mcmahan2017communication}, \textbf{FedBN}~\cite{li2021fedbn} and \textbf{FedRep}~\cite{DBLP:conf/icml/CollinsHMS21} are our competitors. For fair comparison (we should compare different methods with the same scale parameters), we compare our aggregation model $\mathbf{w}_g$ with r50 (ResNet50) backbone to the competitors with r50 backbone, and then use the ensemble model to compare with the competitors with r101 backbone. Note that the original setting of these works is different from ours, they do not have the server-side learning task. Therefore, in the implementation of these methods, we use the server as a client to participate in federated learning.  4) The ideal method: merge the data from the server and all clients together to train an ideal model. Note that the ideal model is not available in real scenarios, we cannot fuse all datasets due to privacy protection policy.

	   \begin{figure}[tb]
		\centering
\subfigure[Our $\mathbf{w}_g$ \textit{vs} competitors-r50]{\includegraphics[width=0.49\linewidth]{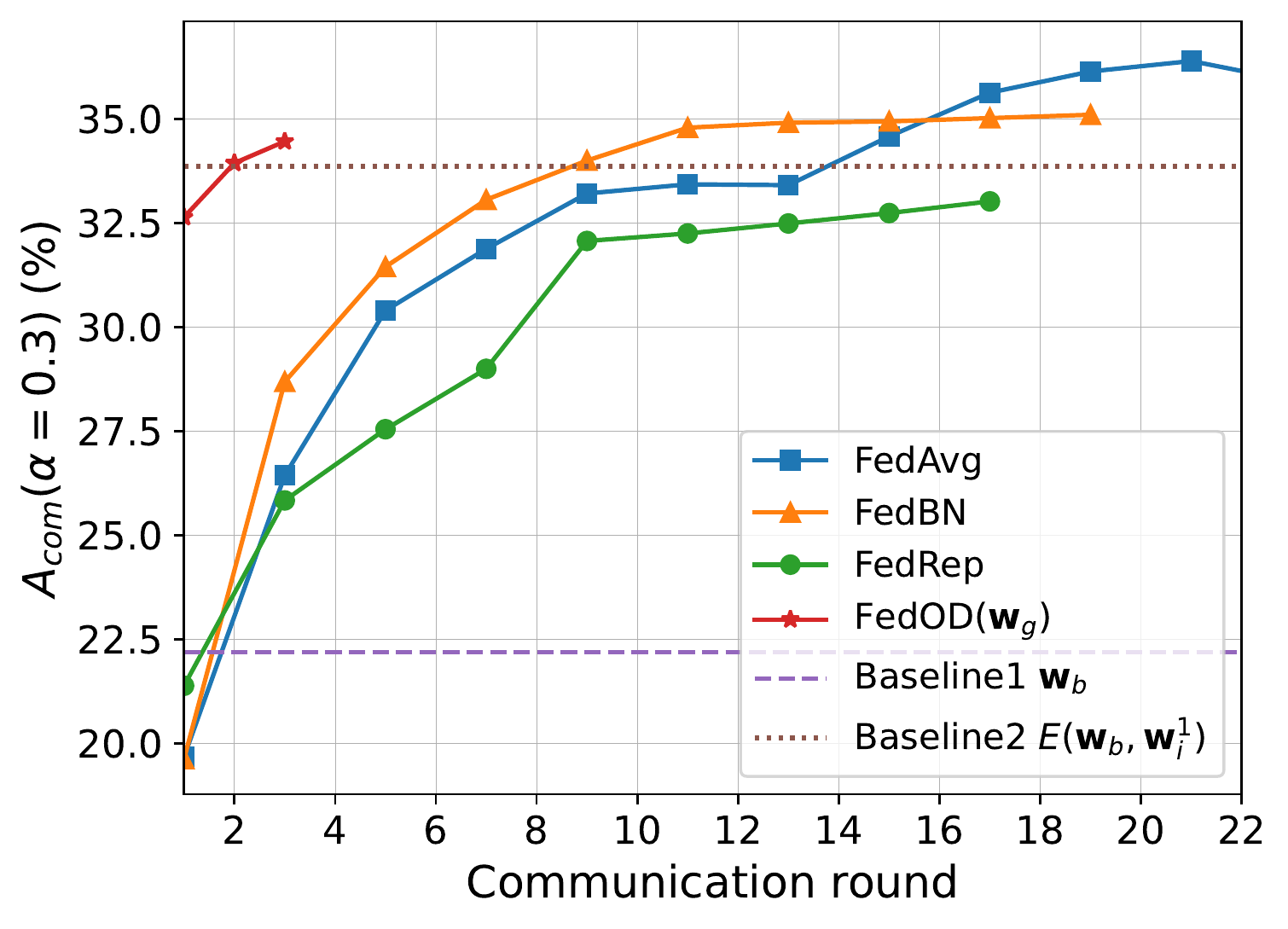}}
\subfigure[Our Ensemble \textit{vs} competitors-r101]{\includegraphics[width=0.49\linewidth]{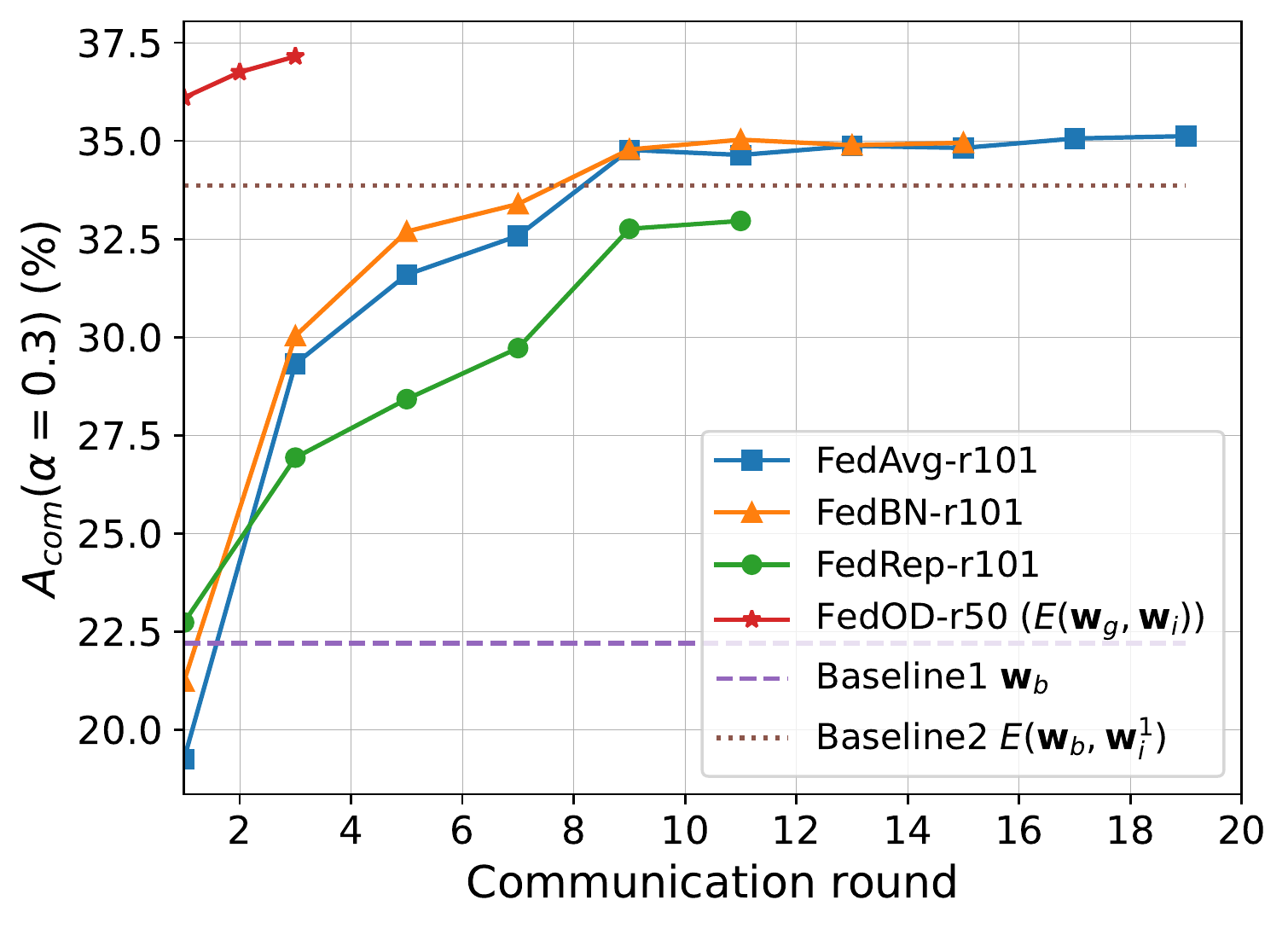}}
		\caption{Performance compared to competitors. FedOD can achieve competitive results with only a few rounds and outperforms the competitors in the global domain.}
		\label{fig:fedavg} 
	\end{figure}
	
	\textbf{Dataset Construction.}  
	We extract data from three public datasets: 1) \textbf{BDD100K}~\cite{yu2020bdd100k} collected in New York, Berkeley, San Francisco and Bay area. It has 100K labeled images with 10 classes for object detection task. 2) \textbf{SODA10M}~\cite{han2021soda10m} collected in 32 cities in China. It consists of 10 million unlabeled images and 20K labeled images. 3) \textbf{NuScenes}~\cite{caesar2020nuscenes} collected in Boston Seaport and Singapore’s One North, Queenstown and Holland Village. It consists of 14 million unlabeled images and 90K labeled images. In addition to regional differences, the data acquisition equipment and the labeling method also produce endogenous differences for the three datasets. Considering the large number of labeled images of BDD100K, we choose it and extract images including cars, pedestrians and riders to build the dataset for the server. Then we use the labeled 20K images in SODA10M to construct two client datasets. These two clients are independent identically distributed, but they are significantly different from the other clients and the server datasets. We select NuScenes to construct another two client datasets. In order to introduce new vehicle instances compared to the other participants, such as trucks and motorcycles, we keep these types of vehicles only on the two NuScenes clients and remove them from the server and the other clients. The overall dataset details are shown in supplementary materials. 
	\subsection{Main Results}
    We present the results of the compared methods. For the federated learning step of FedOD, we set up three rounds of federated learning. The comparison results with Baseline 1 and 2 are shown in Table~\ref{tab:baseline}. We highlight the following points: 
    \begin{itemize}[noitemsep, leftmargin=*]
        \item[1)] \textbf{The base model $\mathbf{w}_b$ cannot handle the client data.} Furthermore, the base model degrades more severely in NuScenes clients than in the two clients built from SODA, reflecting the challenges posed by the imbalance of classes and the unseen instances (trucks, motorcycles, etc.). 
        \item[2)] \textbf{The personalized models $\mathbf{w}_i$ cannot handle the server data.} See $\mathbf{w}^1_i$. $A_p$ increases from 21.15\% to 39.83\%, but $A_s$ drops from 25.5\% to 18.95\% which cause that $\mathbf{w}^1_i$ improves slightly in the union testset, only from 22.60\% to 29.30\%. In comparison, our aggregated model in the 1-th round $\mathbf{w}^1_g$ improves the $A_u$ of $\mathbf{w}_b$ by a large margin, from 22.60\% to 32.50\%. This is because $\mathbf{w}^1_g$ remembers most of the server-side knowledge while improving client-side performance. 
        \item[3)]  \textbf{The ensemble model after federated learning works well.} Both the generalization indicator $A_u$ and the comprehensive indicator $A_{com}$ of our ensemble results outperform the Baseline2. It can be observed that just ensembling the base model $\mathbf{w}_b$ with the fine-tuned client-side model $\mathbf{w}^1_i$ is sub-optimal. $A_{com} (\alpha=0.3)$ can be improved from 33.86\% to 37.15\% following our framework. In addition, with different $\alpha$, the ensemble step of FedOD always achieves the best results. 
        \item[4)] \textbf{Federated learning improves the learning ability of the local model to learn personalized knowledge.} After three rounds of federated learning, the personalized model $\mathbf{w}^4_i$ improves the $A_p$ of $\mathbf{w}^1_i$ from 39.83\% to 42.90\%, even surpassing the ideal model. 
    \end{itemize}
	The proposed FedOD is also compared with the traditional FL method FedAvg~\cite{mcmahan2017communication} and pFL methods, FedBN~\cite{li2021fedbn} and FedRep~\cite{DBLP:conf/icml/CollinsHMS21}. For FedAvg, to be fair, we show the results of our aggregated model $\mathbf{w}_g$ and the ensemble model, respectively. The results are shown in Figure~\ref{fig:fedavg}. It can be seen that the methods within the traditional FL framework converge slowly under the setting of this paper, while FedOD can achieve competitive results with only three rounds of communication. To explore the influence of the amount of parameters on compared methods, we also add the FL methods with ResNet101 backbone as a comparison. We can see our ensemble model outperforms the other compared methods with r101 backbone. By increasing the parameter size, FedAvg-r101 performs better than FedAvg with ResNet50 backbone in the initial stage, but the performance after convergence is mediocre. This observation indicates that simply increasing the parameter size cannot achieve comparable performance to the ensemble. See the supplementary materials for the detailed performance of different methods.



	\begin{figure}[th]
\centering
\resizebox{1\linewidth}{!}{%
    \begin{tabular}{cccc}
        \rotatebox{45}{${GT}$} & \includegraphics[width=.33\linewidth, ]{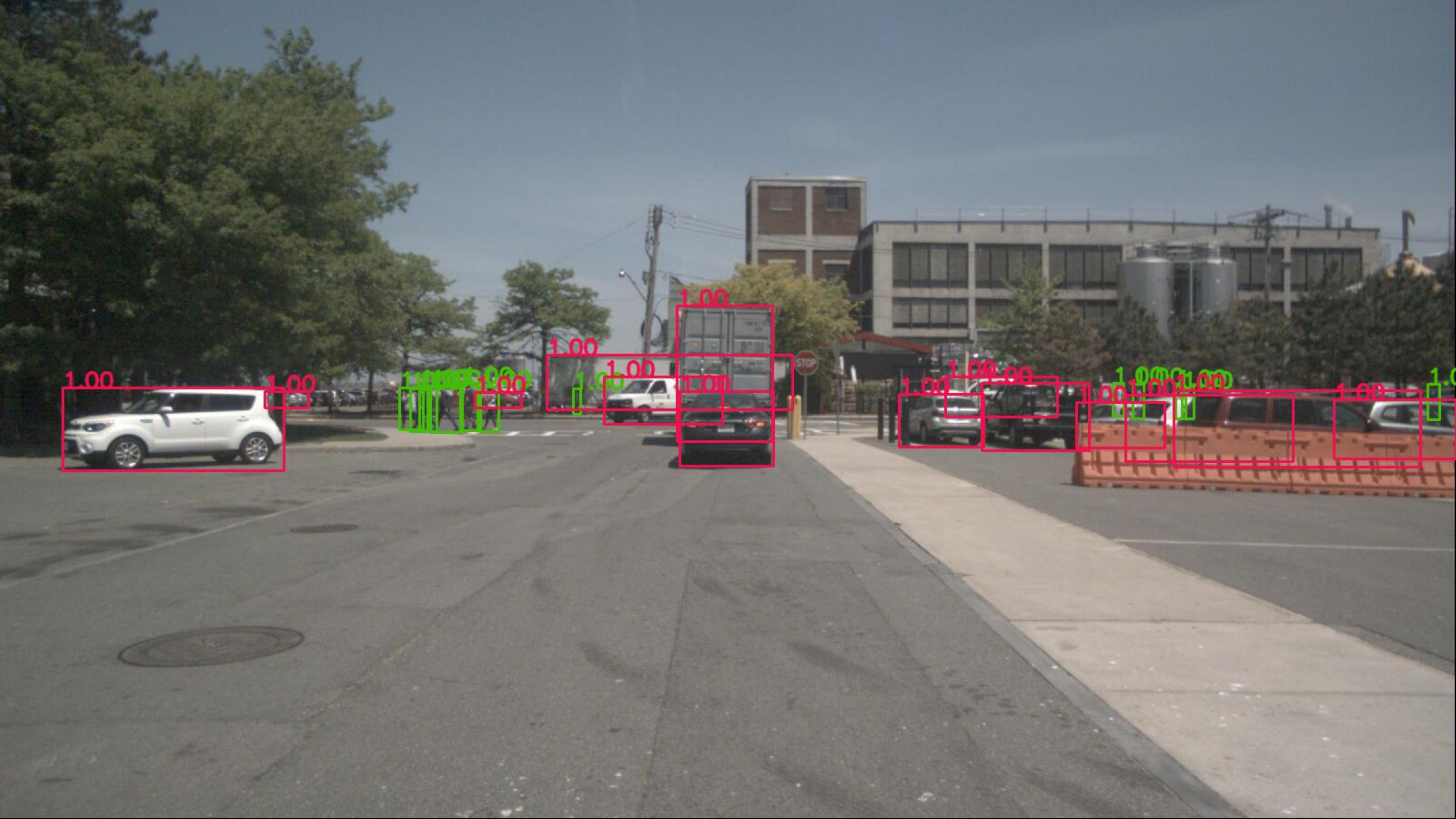} & \includegraphics[width=.33\linewidth, ]{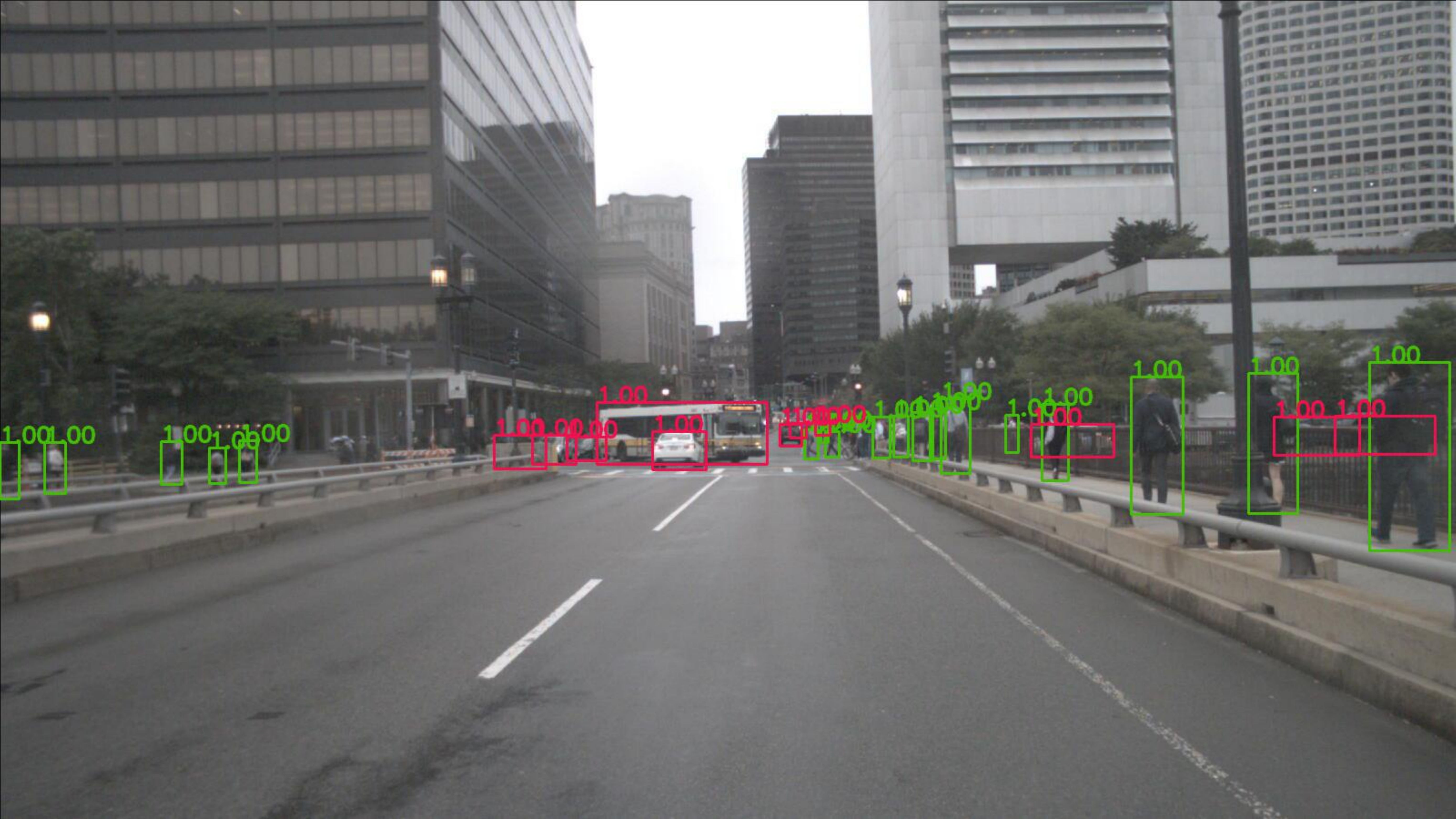} & \includegraphics[width=.33\linewidth, ]{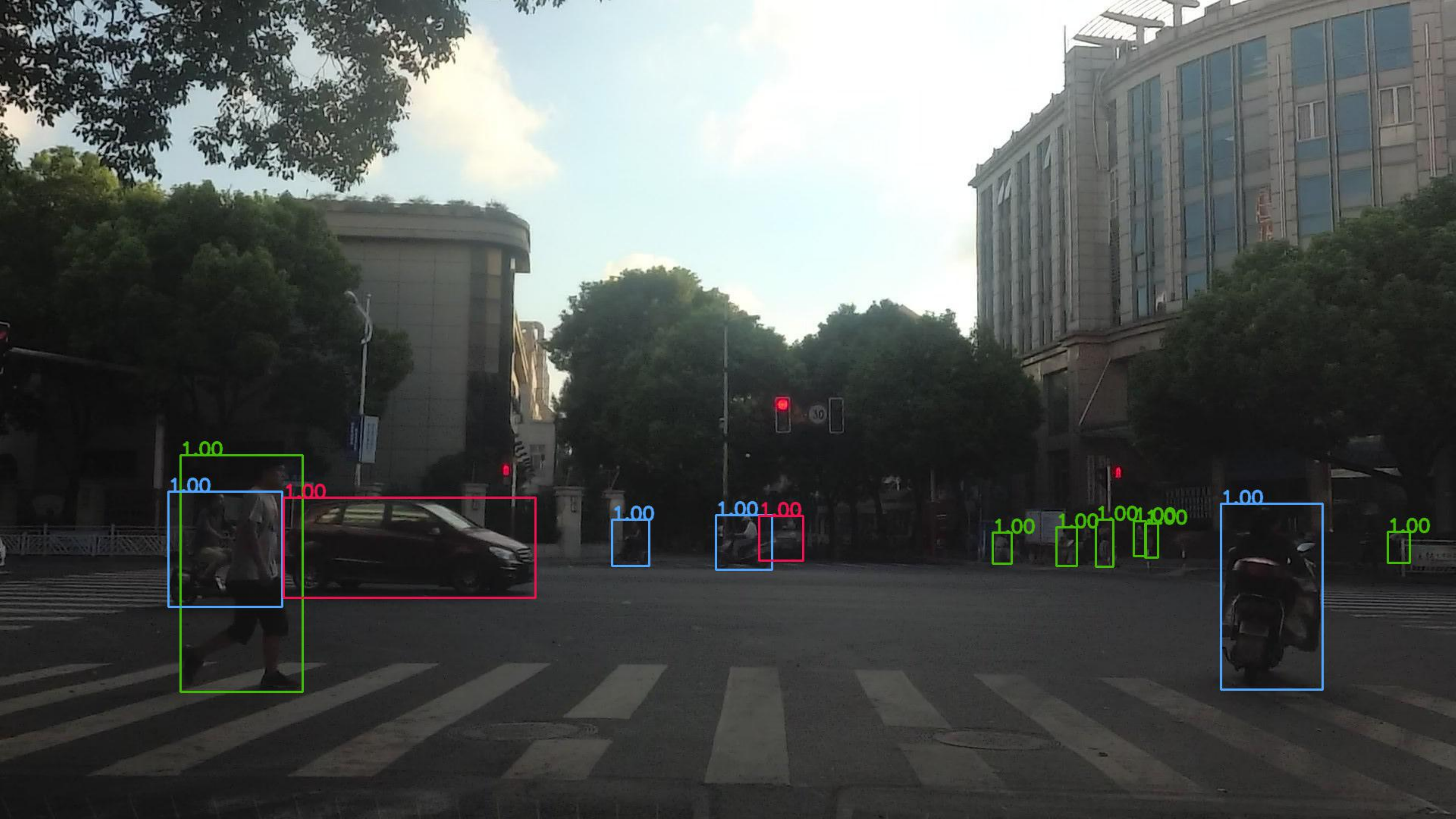}\\
    \rotatebox{45}{${\mathbf{w}_b}$} & \includegraphics[width=.33\linewidth, ]{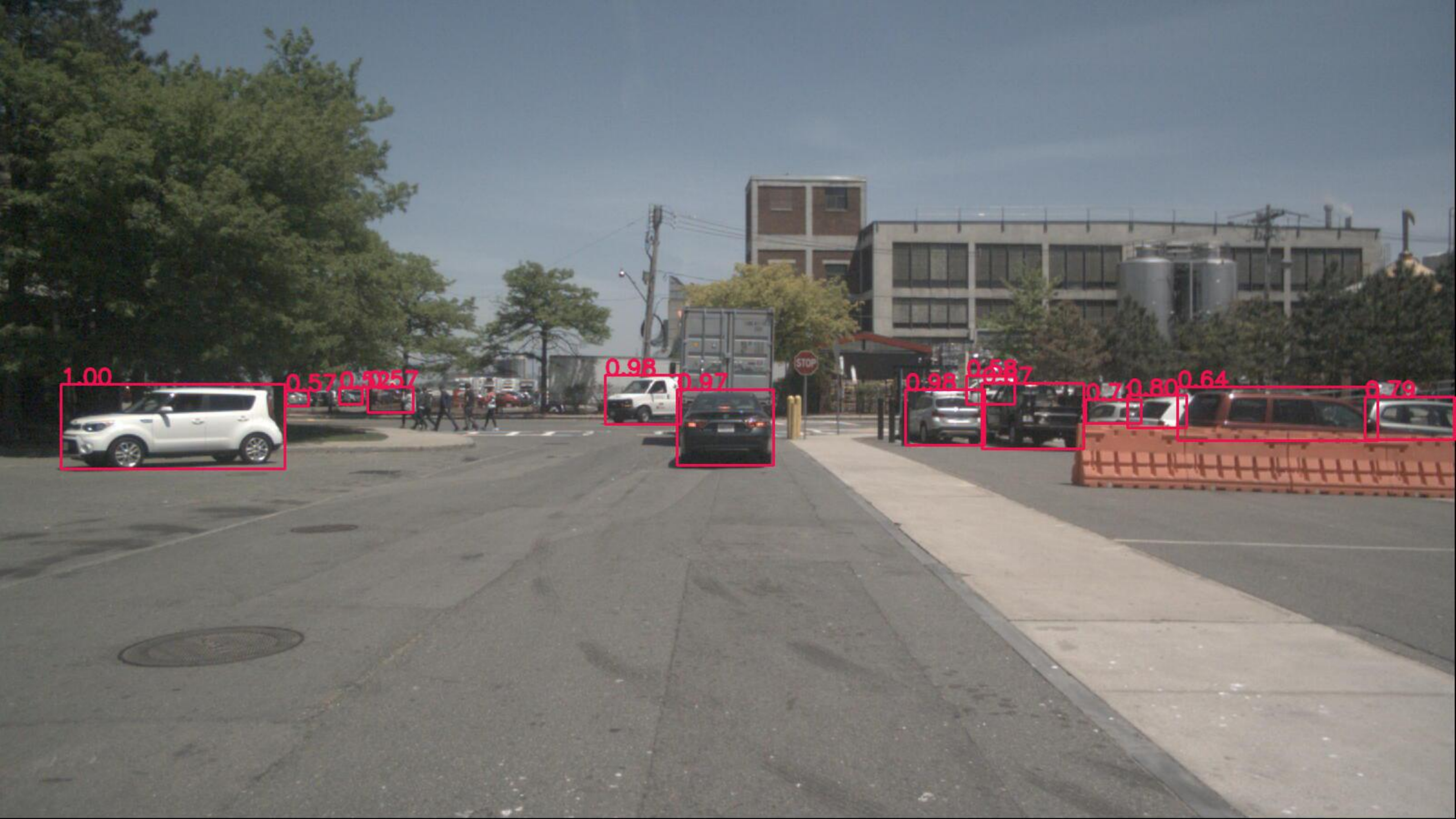} & \includegraphics[width=.33\linewidth, ]{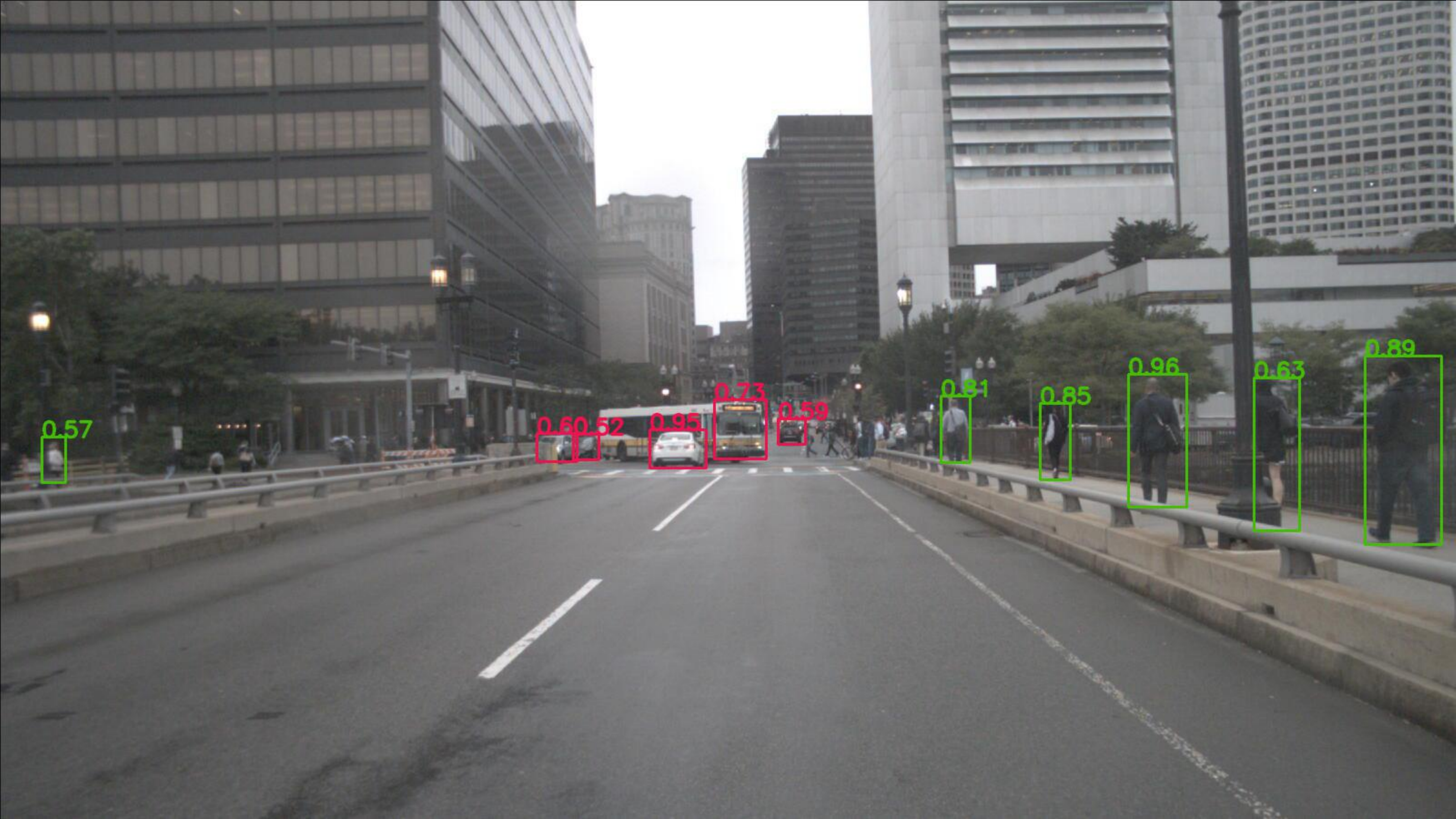} &\includegraphics[width=.33\linewidth, ]{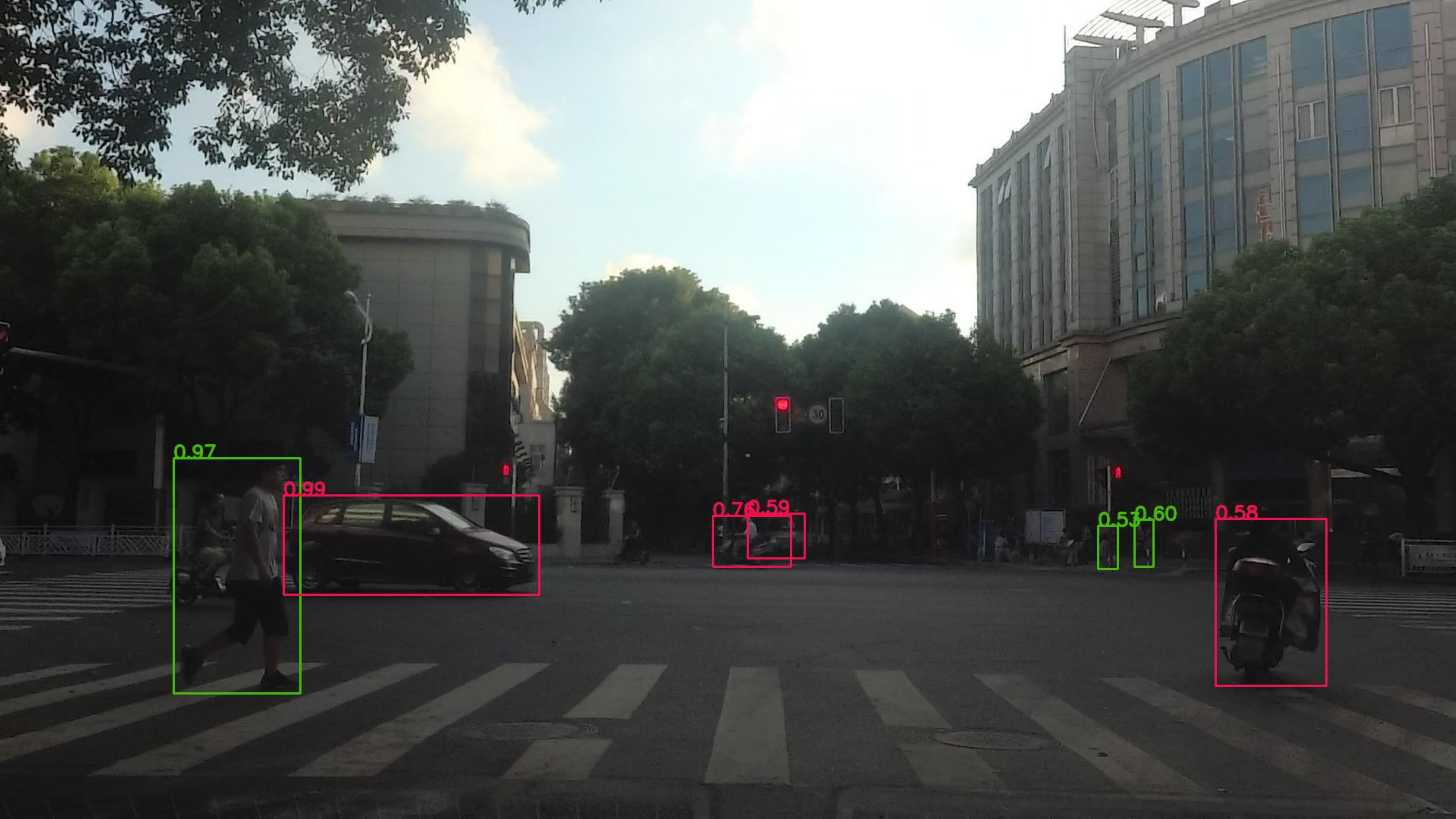}\\
     \rotatebox{45}{$\mathbf{w}^3_3$} & \includegraphics[width=.33\linewidth, ]{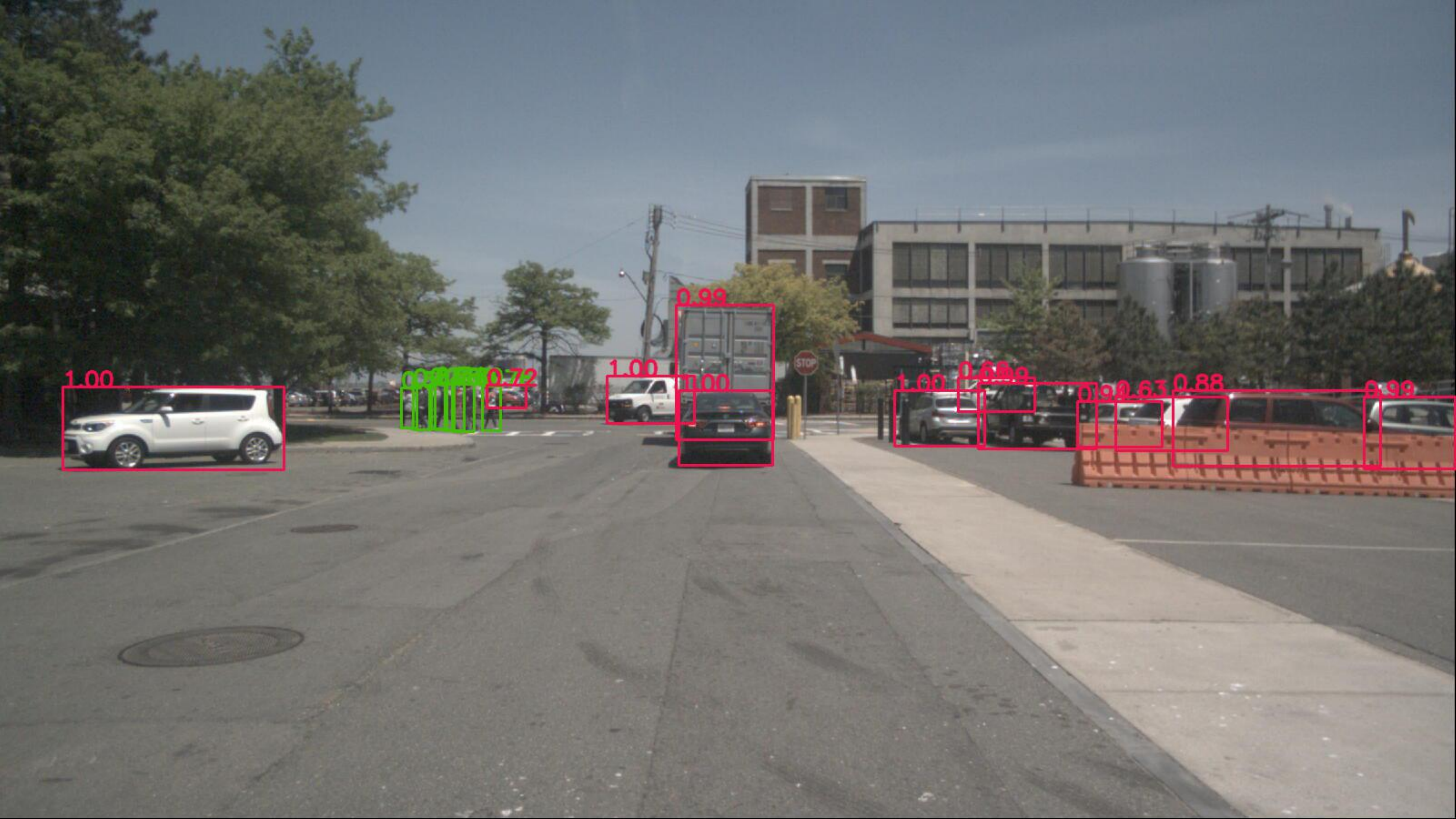} & \includegraphics[width=.33\linewidth, ]{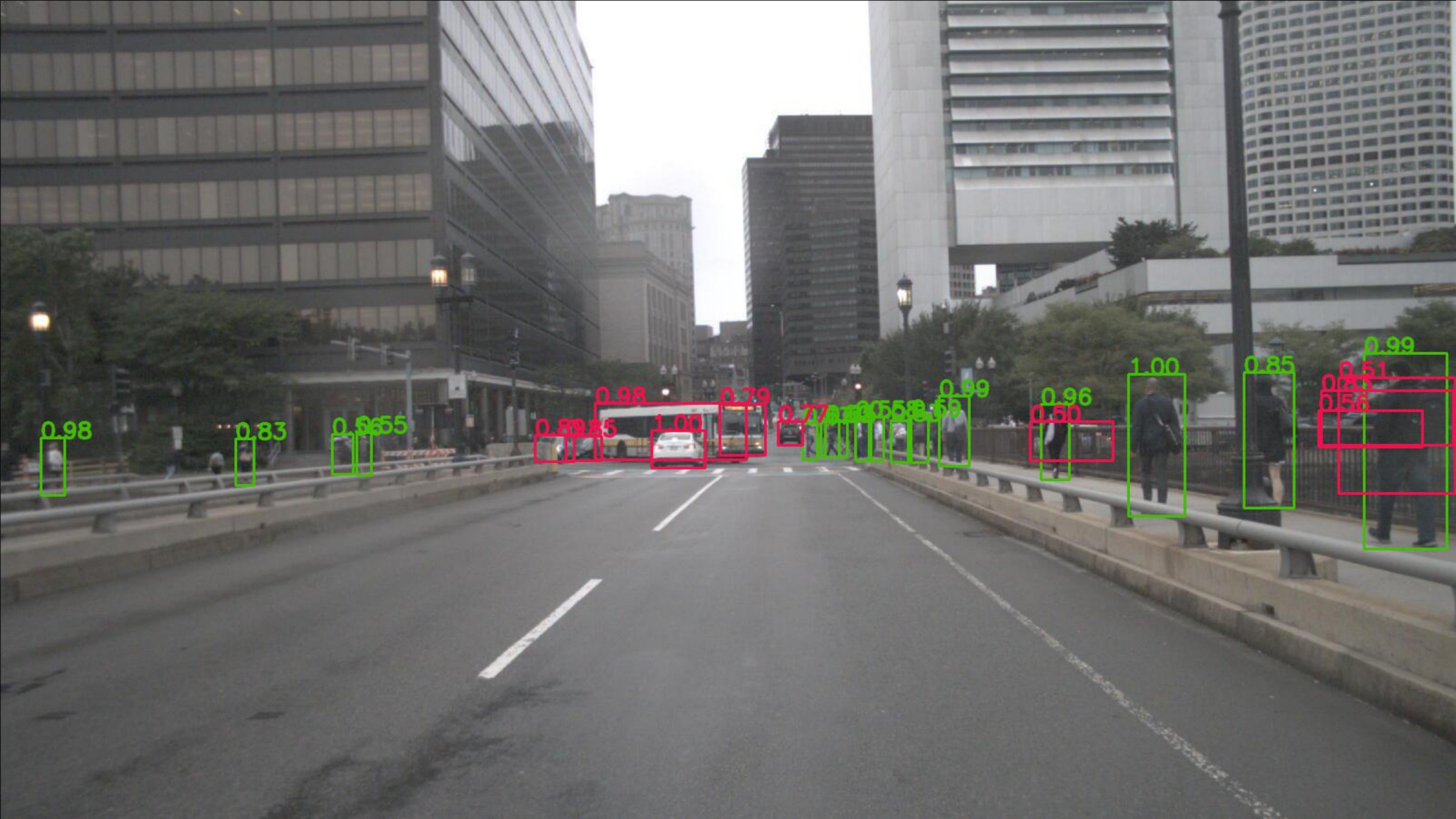} &  \includegraphics[width=.33\linewidth, ]{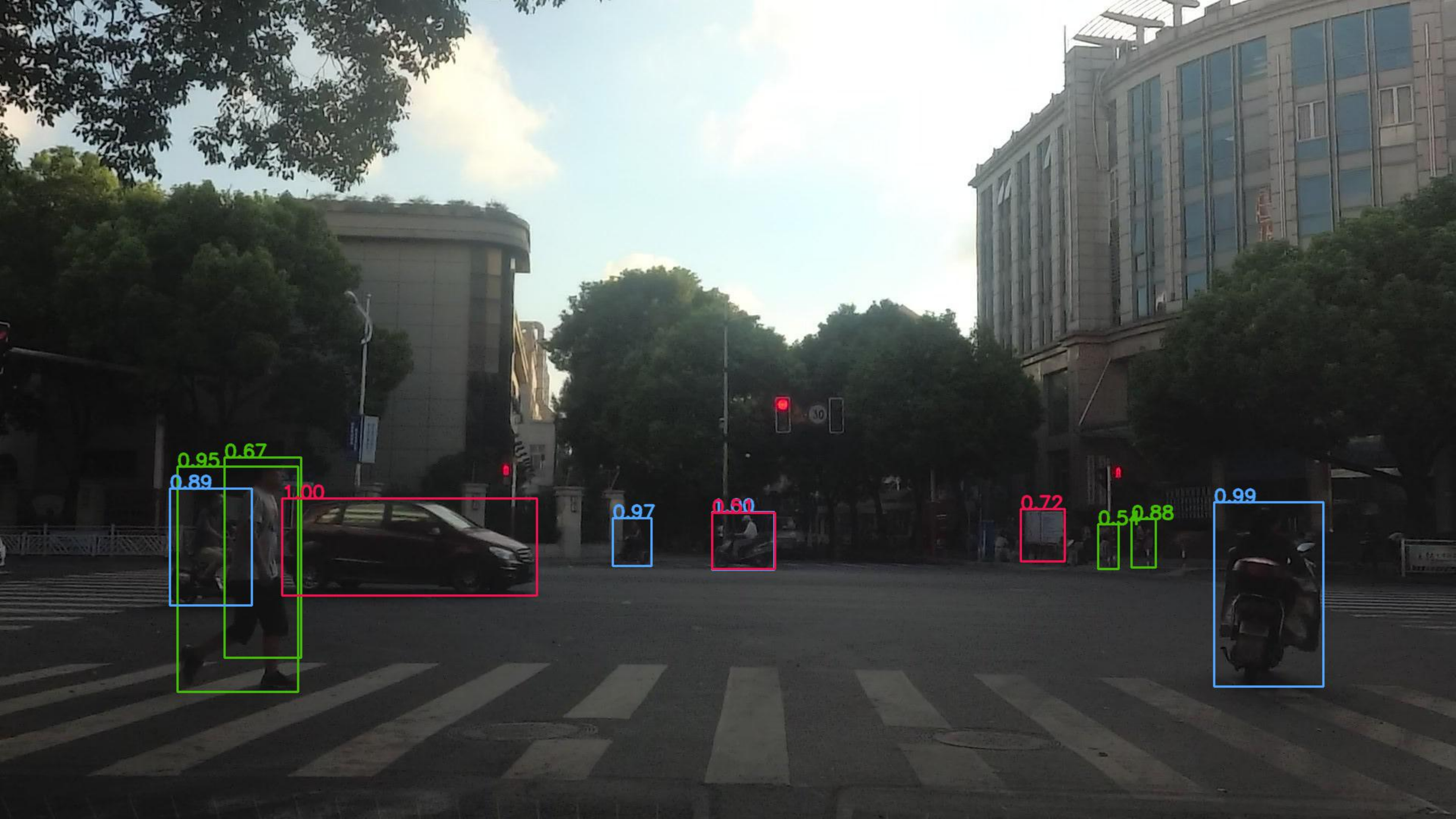}\\
    \rotatebox{45}{$\mathbf{w}^3_g$} & \includegraphics[width=.33\linewidth, ]{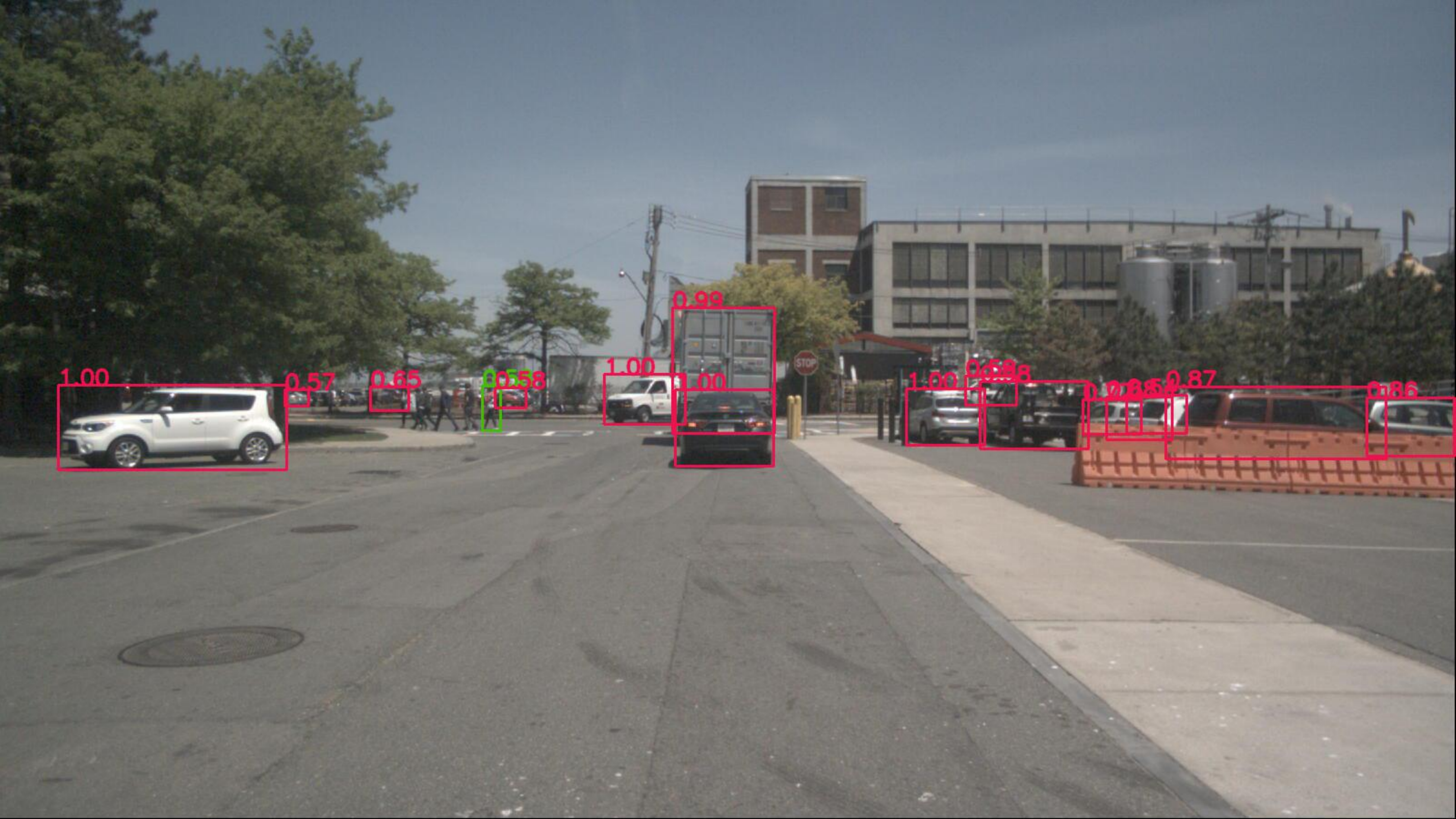} & \includegraphics[width=.33\linewidth, ]{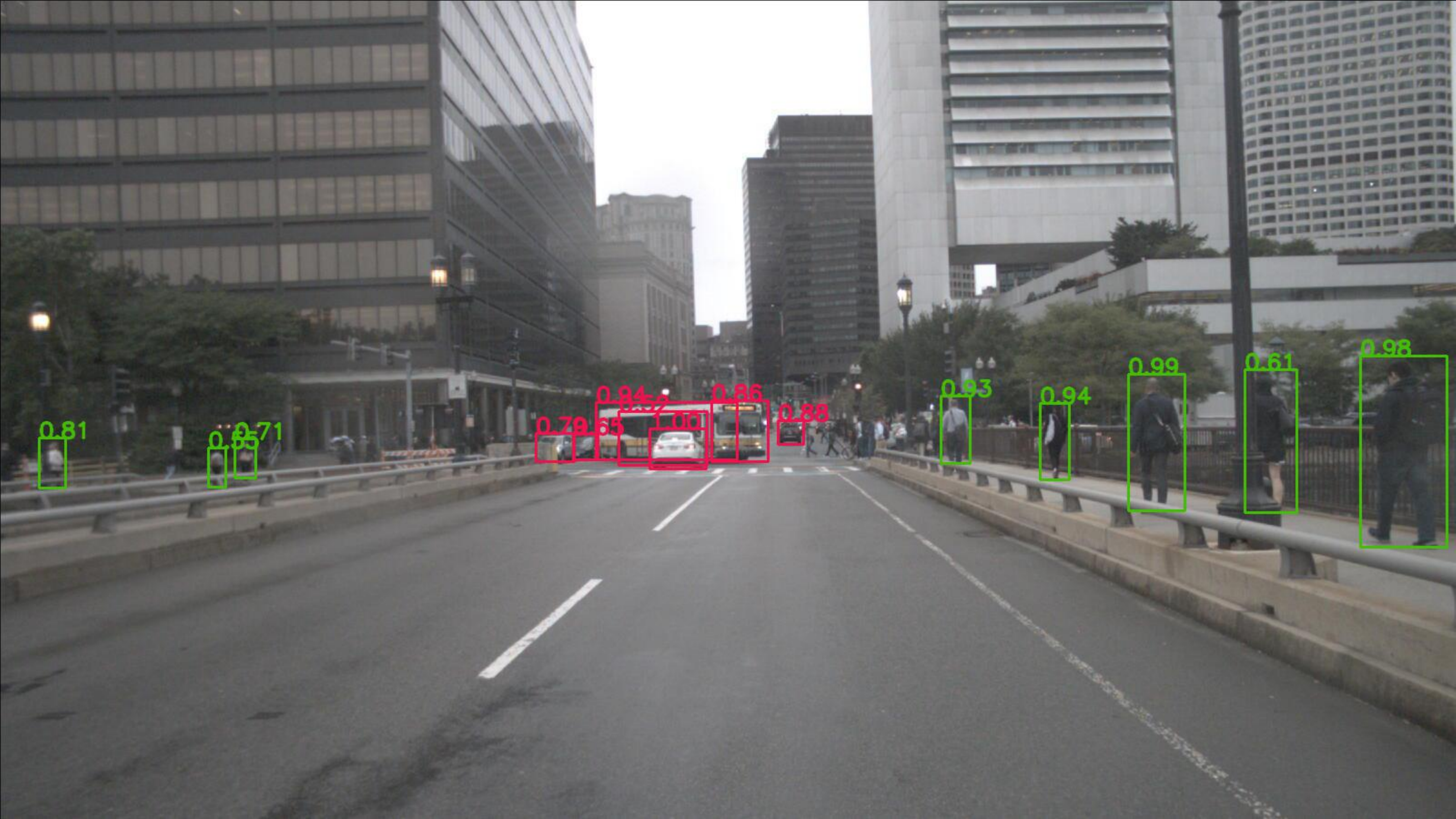} &  \includegraphics[width=.33\linewidth, ]{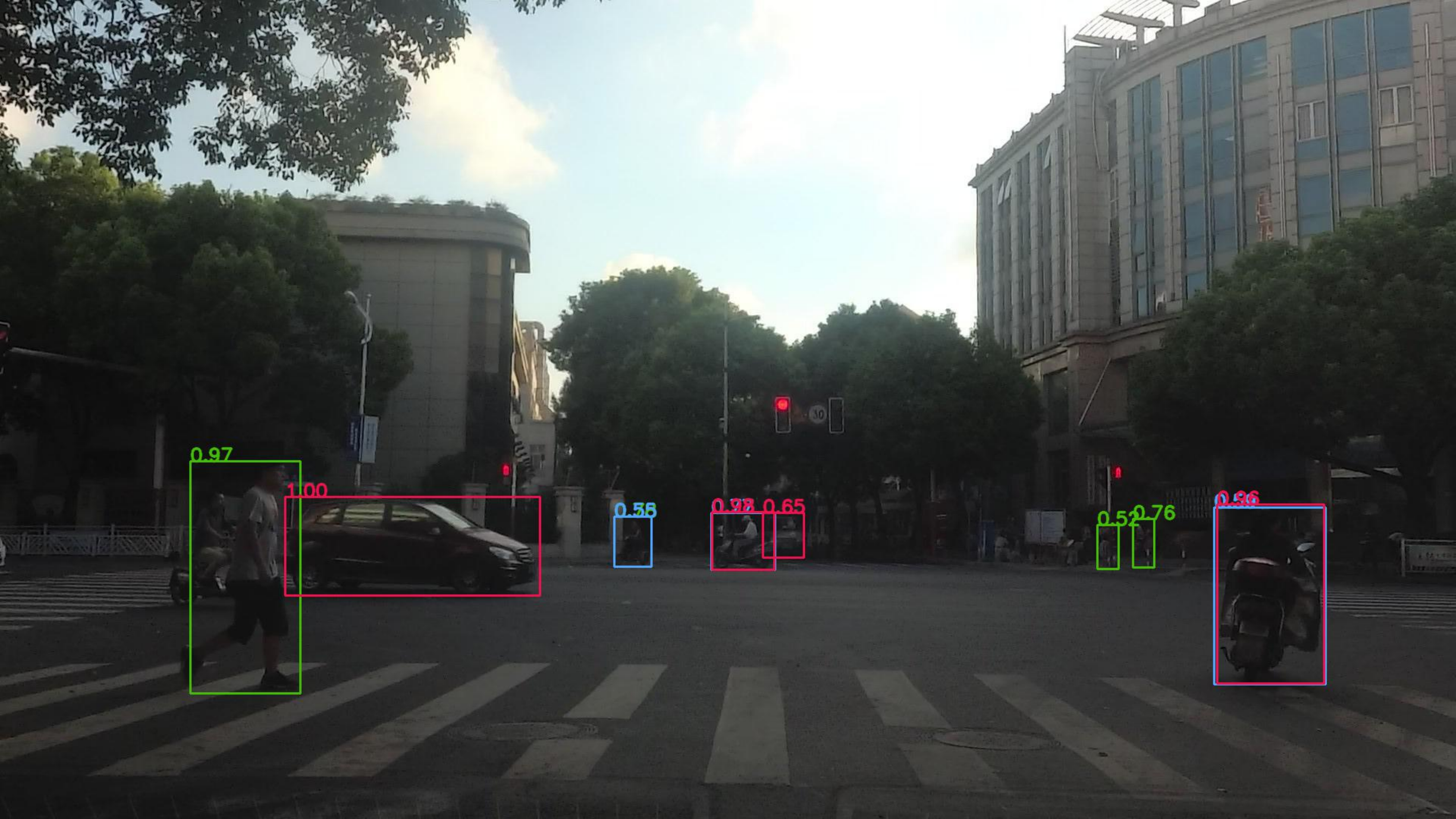}
    \\
    \rotatebox{45}{Ensemble}  & \includegraphics[width=.33\linewidth, ]{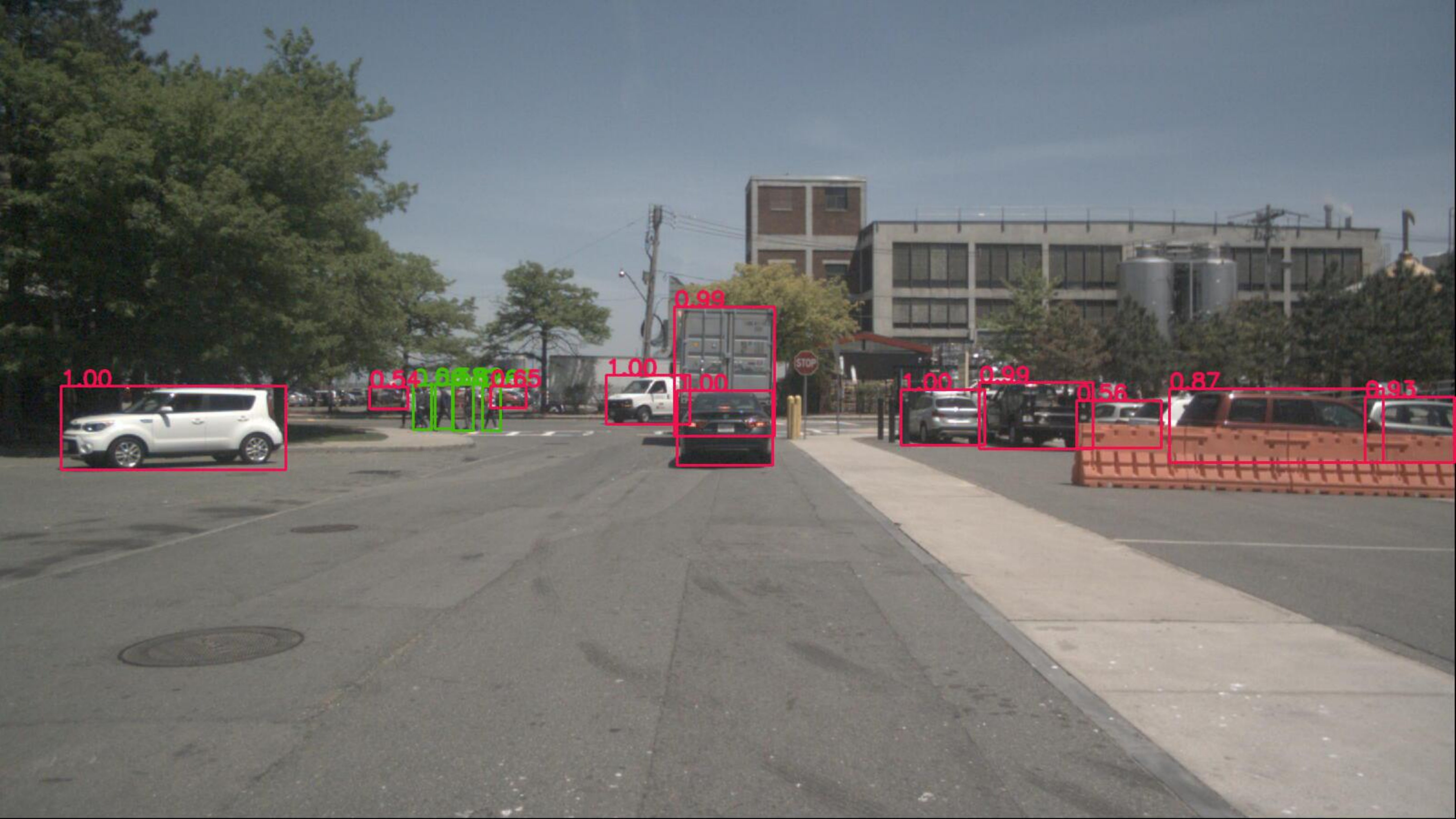} & \includegraphics[width=.33\linewidth, ]{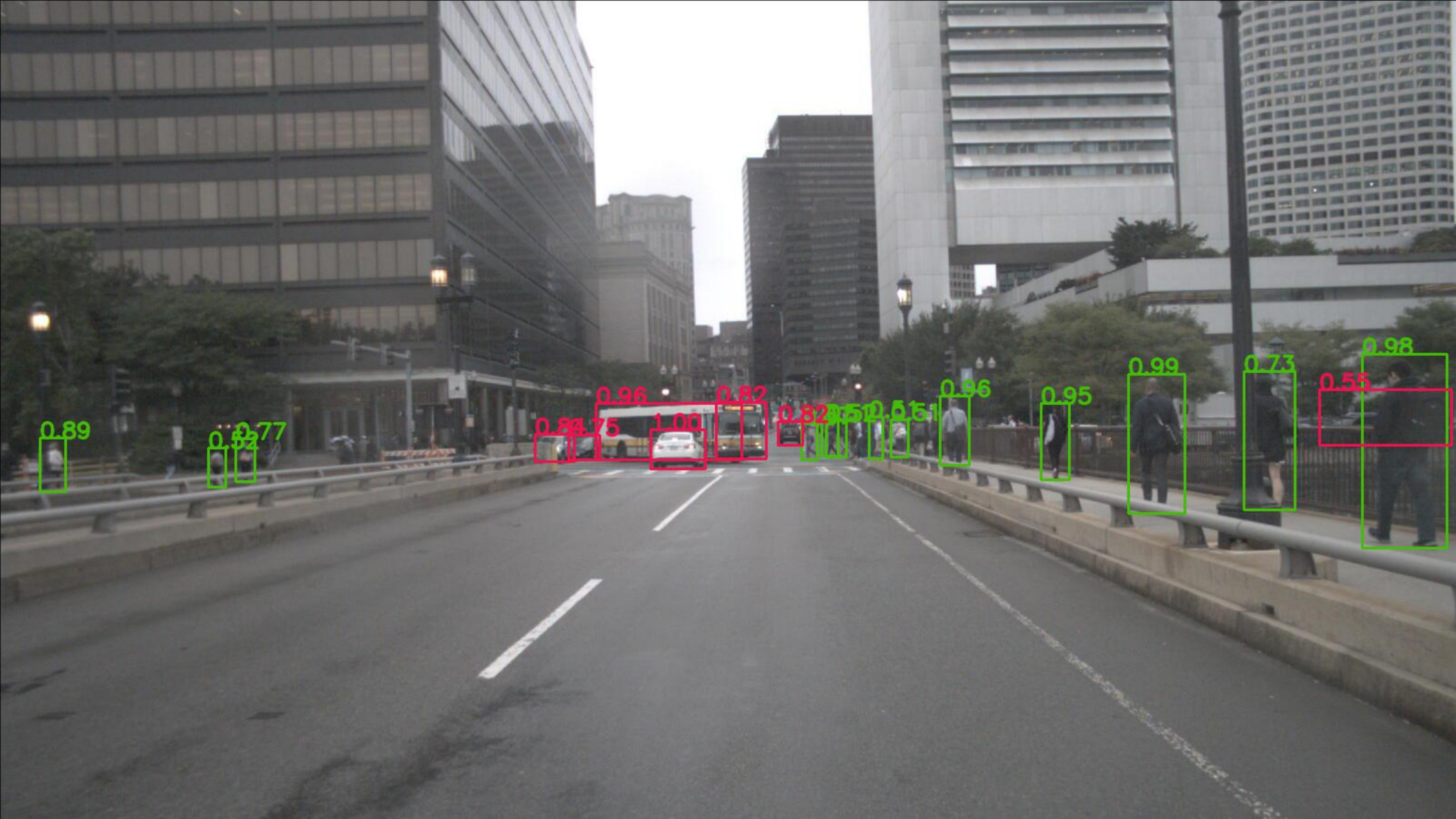} & \includegraphics[width=.33\linewidth, ]{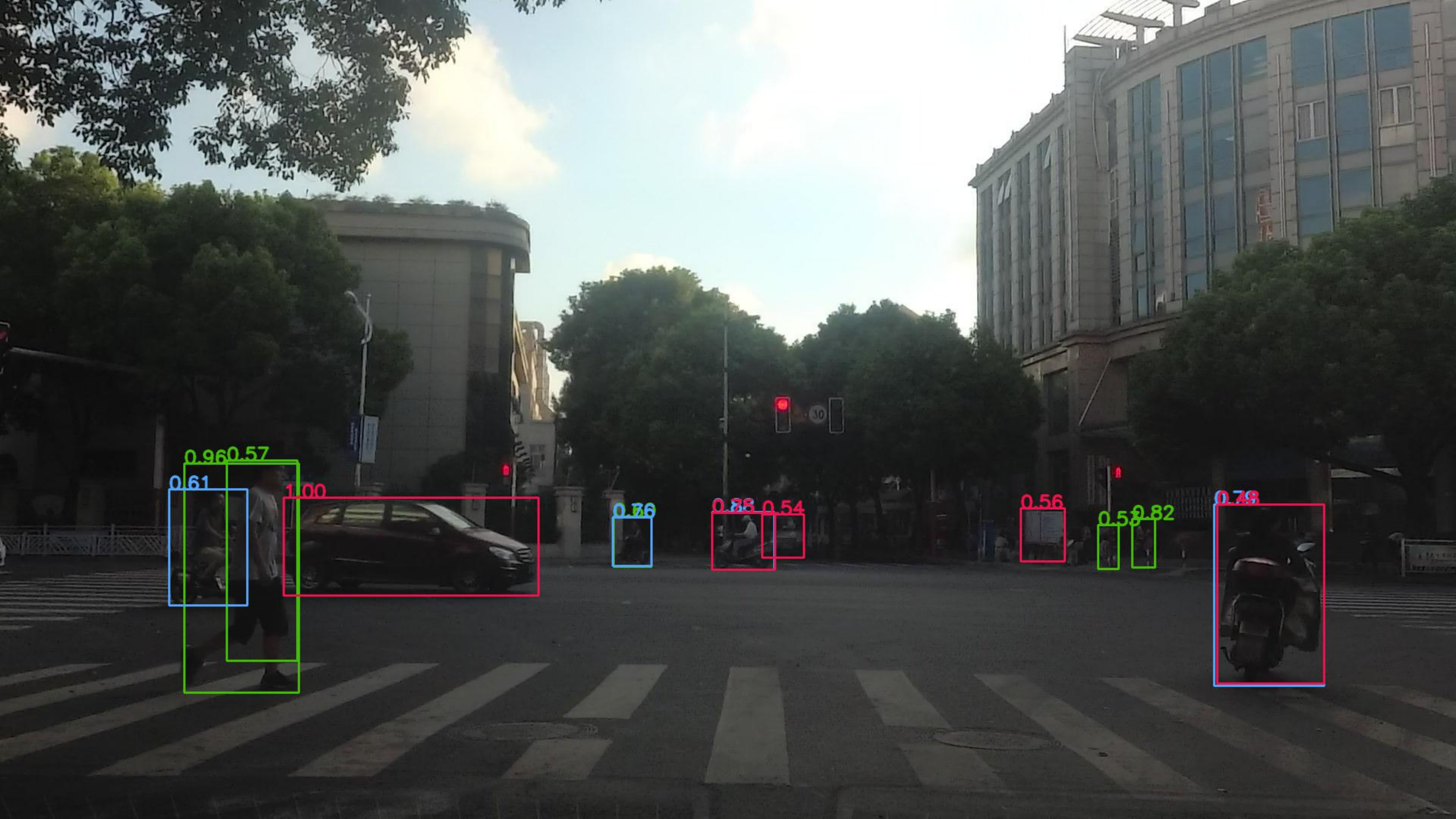}\\

    \end{tabular}}
    \caption{Visualization. The different effects of different models during the federated learning procedure from the perspective of the testset in client 3. The last column is a failure case.
    }
    \label{goodcase}
\end{figure}

	\subsection{Other Results}
\textbf{Visualization Results.} To show the performance changes of different models more concretely during the federated learning, we visualize the detection results in Figure~\ref{goodcase}. The following phenomenons can be observed: 1) The base model $\mathbf{w}_b$ cannot completely detect instances that are not on the server side, such as trucks and buses. 2) Through FedOD, the aggregated model $\mathbf{w}^3_g$ can learn the personalized knowledge of the client. As one can see from the fourth row of the Figure~\ref{goodcase}, the aggregated model $\mathbf{w}^3_g$ is already able to detect trucks and buses, albeit with some redundancy. 

    We also show a failure case in the last column of Figure~\ref{goodcase}, in which, the aggregated model $\mathbf{w}^3_g$ fails to learn the client's personalised knowledge. Due to the severe imbalance in the number of annotations between `Rider' and other classes in the dataset, and there are few instances of `Rider' on each party except client 0 and client 1. Therefore, the information learned by the aggregated model $\mathbf{w}^3_g$ from the client model $\mathbf{w}^3_0$ is limited. This implies that the class imbalance in the federated learning is a direction worthy of further exploration. 
    
\textbf{Different bboxes post-processing methods.} We discuss the impact of different methods to choose the bboxes on the ensemble step. The results are in Table~\ref{tab:nms}.
 
 \begin{table}[tb]
\center
\caption{Comparison of different bboxes filter methods. 
}
\resizebox{0.65\linewidth}{!}{%
\begin{tabular}{cl|ccccl}
\Xcline{1-6}{1pt}
\multicolumn{1}{l}{\textbf{}} & \textbf{} & \multirow{2}{*}{\textbf{NMS}} & \multirow{2}{*}{\textbf{SoftNMS}} & \multirow{2}{*}{\textbf{NWM}} & \multirow{2}{*}{\textbf{WBF}} &  \\
\multicolumn{1}{l}{\textbf{}} & \textbf{} &  &  &  &  & \multicolumn{1}{c}{\textbf{}} \\ \Xcline{1-6}{1pt}
\multicolumn{2}{c|}{$A_s$} & \textit{22.73} & \textit{21.98} & \textit{21.85} & \textit{\textbf{23.13}} &  \\ \cline{1-6}

\multicolumn{2}{c|}{$A_p$} & \textit{39.25} & \textit{39.13} & \textit{37.68} & \textit{\textbf{41.18}} &  \\ \Xcline{1-6}{1pt}
\end{tabular}%
}
\label{tab:nms}
\end{table}

\section{Conclusion}
	This paper proposes a novel cross-domain federated object detection framework FedOD for server-client collaborative autonomous driving scenarios. This framework includes a federated training step based on multi-teacher distillation and an ensemble step based on weighted bboxes fusion. Extensive experiments show that compared to the baselines and competitors, our framework achieves significant performance improvements, and can outperform the traditional FL and pFL methods in all domains.

\section*{Acknowledgment}
This work was supported in part by the National Natural Science Foundation of China (No.62176061), National Key R\&D Program of China (No.2021ZD0112803), STCSM projects (No.20511100400,  No.22511105000), the Shanghai Research and Innovation Functional Program (No.17DZ2260900), and the Program for Professor of Special Appointment (Eastern Scholar) at Shanghai Institutions of Higher Learning.

\small{
\bibliographystyle{IEEEbib}
\bibliography{icme2023}}

\end{document}